\PassOptionsToPackage{prologue,dvipsnames}{xcolor}
\documentclass[sigconf]{acmart}
\AtBeginDocument{%
  }

\usepackage{soul}
\usepackage{url}
\usepackage[utf8]{inputenc}
\usepackage{graphicx}
\usepackage{amsmath}
\usepackage{amsthm}
\usepackage{booktabs}
\usepackage{algorithm}
\usepackage{algorithmic}
\usepackage{multirow}
\usepackage{multicol}
\usepackage{bm}
\usepackage{underscore}
\usepackage{amsfonts}
\usepackage{xcolor}
\usepackage{color, colortbl}
\usepackage{makecell}
\usepackage{enumitem}

\usepackage{amssymb}

\usepackage{colortbl}

\usepackage[switch]{lineno}

\definecolor{mygray}{gray}{.9}
\definecolor{gray2}{gray}{.78}
\definecolor{gray3}{gray}{.7}
\definecolor{gray4}{gray}{.6}
\definecolor{gray5}{gray}{.5}
\newcommand{\CC}{\cellcolor{mygray}}
\newcommand{\CK}{\cellcolor{gray2}}

\newcommand{\rmnum}[1]{\romannumeral #1}
\newcommand{\Rmnum}[1]{\expandafter\@slowromancap\romannumeral #1@}

\newcommand{\wen}[1]{{\color{black}#1}}

\copyrightyear{2023}
\acmYear{2023}
\setcopyright{acmlicensed}
\acmConference[MM '23] {Proceedings of the 31st ACM International Conference on Multimedia}{October 29--November 3, 2023}{Ottawa, ON, Canada.}
\acmBooktitle{Proceedings of the 31st ACM International Conference on Multimedia (MM '23), October 29--November 3, 2023, Ottawa, ON, Canada}
\acmPrice{15.00}
\acmISBN{979-8-4007-0108-5/23/10}
\acmDOI{10.1145/3581783.3611786}
\begin{document}

\title{MEAformer: Multi-modal Entity Alignment Transformer \\ for Meta Modality Hybrid}




\author{Zhuo Chen}
\orcid{0000-0001-9991-6892}
\affiliation{%
  \institution{Zhejiang University}
  \city{Hangzhou}
  \state{Zhejiang}
  \country{China}
  }
\email{zhuo.chen@zju.edu.cn}

\author{Jiaoyan Chen}
\orcid{0000-0003-4643-6750}
\affiliation{%
  \institution{The University of Manchester}
  \city{Manchester}
  \country{United Kingdom}
}
\email{jiaoyan.chen@manchester.ac.uk}

\author{Wen Zhang}
\orcid{0000-0001-8429-9326}
\authornote{corresponding author.}
\affiliation{%
 \institution{Zhejiang University}
  \city{Ningbo}
  \state{Zhejiang}
  \country{China}
  }
\email{zhang.wen@zju.edu.cn}

\author{Lingbing Guo, Yin Fang}
\affiliation{%
  \institution{Zhejiang University}
  \city{Hangzhou}
  \state{Zhejiang}
  \country{China}
  }
\email{{lbguo,fangyin}@zju.edu.cn}

\author{Yufeng Huang, Yichi Zhang}
\affiliation{%
  \institution{Zhejiang University}
  \city{Hangzhou}
  \state{Zhejiang}
  \country{China}
  }
\email{{huangyufeng,zhangyichi2022}@zju.edu.cn}

\author{Yuxia Geng}
\orcid{0000-0002-2461-2613}
\affiliation{%
  \institution{Zhejiang University}
  \city{Hangzhou}
  \state{Zhejiang}
  \country{China}
  }
\email{gengyx@zju.edu.cn}

\author{Jeff Z. Pan}
\orcid{0000-0002-9779-2088}
\affiliation{%
  \institution{The University of Edinburgh}
  \city{Edinburgh}
  \country{United Kingdom}
}
\email{j.z.pan@ed.ac.uk}

\author{Wenting Song}
\orcid{0000-0002-3163-2399}
\affiliation{
  \institution{Huawei Technologies Ltd}
  \city{Xi'an}
  \state{Shanxi}
  \country{China}
  }
\email{songwenting@huawei.com}

\author{Huajun Chen}
\orcid{0000-0001-5496-7442}
\affiliation{%
  \institution{Zhejiang University \\ Donghai laboratory}
  \city{Hangzhou}
  \state{Zhejiang}
  \country{China}
}
\email{huajunsir@zju.edu.cn}

\renewcommand{\shortauthors}{Chen et al.}

\begin{abstract}
Multi-modal entity alignment (MMEA) aims to discover identical entities across different knowledge graphs (KGs) whose entities are associated with relevant images.
However, current MMEA algorithms rely on KG-level modality fusion strategies for multi-modal entity representation, which ignores the variations of modality preferences of different entities, thus compromising robustness against noise in modalities such as blurry images and relations. 
This paper introduces MEAformer, a \textbf{m}ulti-modal \textbf{e}ntity \textbf{a}lignment trans\textbf{former} approach for meta modality hybrid, which dynamically predicts the mutual correlation coefficients among modalities for more fine-grained entity-level modality fusion and alignment. 
Experimental results demonstrate that our model not only achieves SOTA performance in multiple training scenarios, including supervised, unsupervised, iterative, and low-resource settings, but also has a limited number of parameters, efficient runtime, and interpretability.  
Our code is available at {\color{NavyBlue} \url{github.com/zjukg/MEAformer}}. 

\end{abstract}




\maketitle

\vspace{-2pt}
\begin{figure}[!htbp]
  \centering
   \vspace{-3pt}
\includegraphics[width=0.91\linewidth]{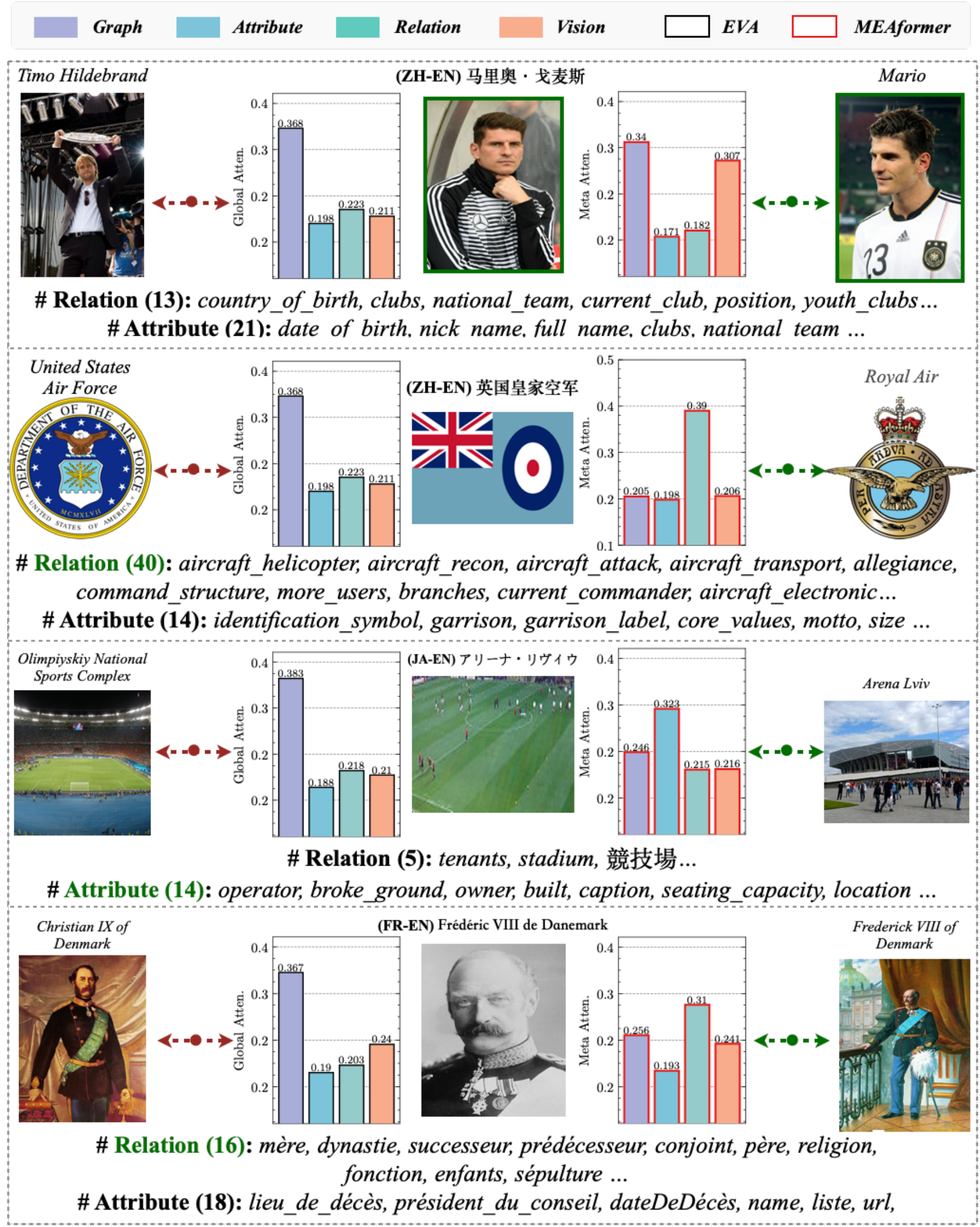}
  \vspace{-3pt}
  \caption{Static modality weights in EVA~\protect\cite{DBLP:conf/aaai/0001CRC21} (left) and dynamic meta modality weights in our MEAformer (right).
  }
  \label{fig:Introcase}
  \vspace{-8pt}
\end{figure}
\section{Introduction}
In recent years, knowledge graphs (KGs) have increasingly facilitated numerous AI applications such as question answering \cite{DBLP:conf/semweb/0007CGPYC21,DBLP:conf/jist/0007HCGFP0Z22} 
and AI4Science \cite{fang2023knowledge,DBLP:conf/aaai/FangZYZD0Q0FC22}, providing commonsense knowledge. 
As a critical task in KG integration and construction, entity alignment (EA) 
aims to identify equivalent entities across KGs while addressing challenges such as disparate naming conventions, multilingualism, and heterogeneous graph structures.  
To leverage visual content from the Internet as supplementary information for EA, multi-modal entity alignment (MMEA) has been proposed, wherein each entity is associated with its name-related images \cite{DBLP:conf/ksem/ChenLWXWC20}.
Current MMEA approaches primarily concentrate on devising an appropriate cross-KG modality fusion paradigm.
\wen{\citeauthor{DBLP:conf/aaai/0001CRC21}} \cite{DBLP:conf/aaai/0001CRC21}
introduce modality-specific attention weight learning for modality importance; 
\wen{\citeauthor{DBLP:conf/kdd/ChenL00WYC22}} \cite{DBLP:conf/kdd/ChenL00WYC22} integrate visual features to guide the relation and attribute learning;
\wen{\citeauthor{DBLP:conf/coling/LinZWSW022}} \cite{DBLP:conf/coling/LinZWSW022} apply KL divergence over the output distribution between joint  and uni-modal entity embedding to reduce the modality gap.
However, they all learn KG-level weights for modality fusion, disregarding intra-modal discrepancies (e.g., node degrees or relation numbers) and inter-modal preferences (e.g., modality absence or ambiguity) for each entity. 
These flaws somehow impact their robustness.

In this work, we explore an alternative MMEA paradigm that generates reasonable multi-modal entity hybrid features to foster modality preferences adaptable to entities. 
Specifically, we propose a novel \textbf{m}ulti-modal \textbf{e}ntity \textbf{a}lignment trans\textbf{former} approach named MEAformer, 
which dynamically predicts relatively mutual weights among modalities for each entity (see examples in Figure \ref{fig:Introcase}).
We implement our model as a meta-learning \cite{DBLP:journals/pami/HospedalesAMS22} alike paradigm, where our proposed 
dynamic cross-modal weighted (DCMW) module generates the meta modality weights for modality correction,
enabling inter-modal mutual rating through a shallow cross-attention network.
Furthermore,  we employ a modal-adaptive contrastive learning objective to efficiently disentangle modality information with limited 
pre-aligned entities,
and introduce a modal-aware hard entity replay to further augment \wen{the} model's robustness for vague entity details.
Overall, our contributions can be summarized as:
\begin{itemize}
 \item We identify the limitations of the KG-level modality fusion strategies in existing MMEA research, and propose to use a more adaptive entity-level modality fusion strategy.
 \wen{\item We develop an effective transformer-based MMEA method called MEAformer, which can dynamically predict weights between modalities via utilizing techniques such as meta-learning and contrastive learning for implementation.
 }
 \item 
 \wen{We conduct experiments demonstrating}
 that MEAformer is able to achieve SOTA performance across multiple training scenarios 
with simple framework, limited parameters, optimistic speed, and strong interpretability.
\end{itemize}

\vspace{-1pt}
\section{Related Work}
\begin{figure*}[!htbp]
  \centering
  \includegraphics[width=0.99\linewidth]{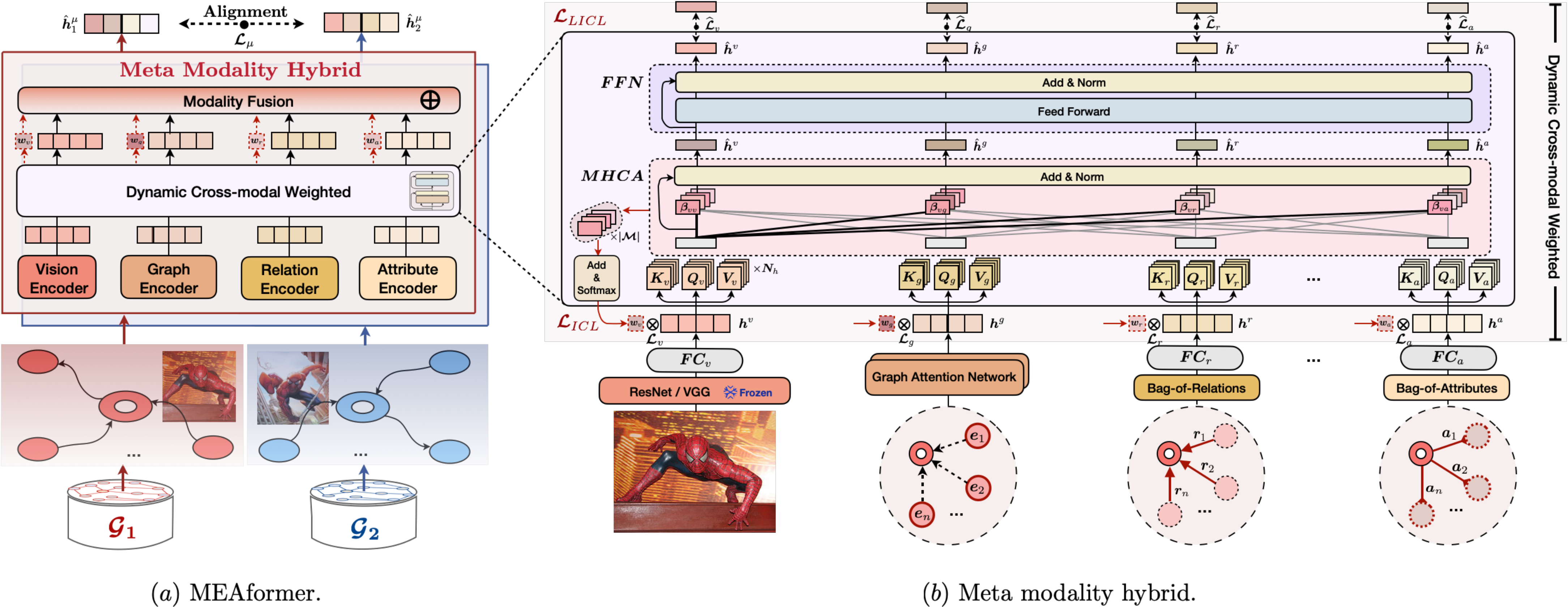}
  \vspace{-7pt}
  \caption{(a) The overall framework of MEAformer and (b) the implementation details of meta modality hybrid (MMH).  
  }
 \vspace{-4pt}
  \label{fig:model}
\end{figure*}
\subsection{Entity Alignment}
Entity Alignment (EA) aims to discover equivalent entities across different KGs to facilitate knowledge fusion.
Early EA systems exploited techniques such as logical reasoning and lexical matching for entity mapping construction \cite{DBLP:conf/semweb/Jimenez-RuizG11,DBLP:journals/pvldb/SuchanekAS11}, heavily relying on (ad hoc) heuristics.
Recent embedding-based EA methods alleviate the heterogeneity issue \cite{DBLP:journals/tkde/SunHWWQ23} by learning an embedding space to represent those to-be-aligned KGs, where similar entities are closely situated while dissimilar ones are separated far apart. 
Concretely, we categorize  them into two categories \cite{DBLP:journals/vldb/ZhangTLJQ22}:
\textbf{\textit{(\rmnum{1})}} {\emph{GNNs-based EA methods}} \cite{DBLP:conf/aaai/SunW0CDZQ20,DBLP:conf/emnlp/LiuCPLC20,DBLP:conf/acl/WuLFWZ20,DBLP:conf/kdd/GaoLW0W022} mainly utilize graph neural networks (GNNs)  like GCN \cite{DBLP:conf/iclr/KipfW17} and GAT \cite{DBLP:conf/iclr/VelickovicCCRLB18} for neighborhood entity feature aggregation.
\textbf{\textit{(\rmnum{2})}} {\emph{Translation-based EA methods}} \cite{DBLP:conf/ijcai/ZhangSHCGQ19,DBLP:conf/semweb/SunHHCGQ19,DBLP:conf/wsdm/XinSH0022,DBLP:conf/ijcai/CaiMZJ22} 
adopt translation-based KG embedding methods like TransE \cite{DBLP:conf/nips/BordesUGWY13} 
to capture the entity structure information from relational triples.

Commonly, an alignment objective (e.g., embedding cosine similarity) is applied on part of the pre-aligned entity pairs (a.k.a. seed alignments) to calibrate KGs' semantic space in these methods.
Previous works demonstrated that the EA progress could be enhanced through additional strategies, such as: parameter sharing \cite{DBLP:conf/ijcai/ZhuXLS17} (i.e., sharing the entity embedding of seed alignments across KGs); explicitly linking the seed alignments among multiple heterogeneous KGs \cite{DBLP:conf/ijcai/ZhuXLS17};  iterative learning \cite{DBLP:conf/ijcai/SunHZQ18}  (i.e., iteratively labeling entity pairs as pseudo seed supervision); attribute value encoding \cite{DBLP:conf/aaai/TrisedyaQZ19}; collective stable matching for interdependent alignment decisions \cite{DBLP:conf/icde/Zeng0T020}; or guiding EA through ontological schemas \cite{DBLP:conf/acl/XiangZCCLZ21}.

 \subsection{Multi-modal Entity Alignment}
Since being introduced by \citeauthor{DBLP:conf/esws/LiuLGNOR19} \cite{DBLP:conf/esws/LiuLGNOR19} as a task of the multi-modal knowledge graph (MMKG) construction, incorporating the visual modality for EA in KGs has gradually gained attention in communities with the development of multi-modal learning in recent years.
\citeauthor{DBLP:conf/ksem/ChenLWXWC20} \cite{DBLP:conf/ksem/ChenLWXWC20} fuse the knowledge representations of modalities and then minimize the distance between the holistic embeddings of aligned entities. \citeauthor{DBLP:conf/aaai/0001CRC21} \cite{DBLP:conf/aaai/0001CRC21} apply a learnable attention weighting scheme to  give varying importance to each modality. 
\citeauthor{DBLP:conf/kdd/ChenL00WYC22} \cite{DBLP:conf/kdd/ChenL00WYC22} integrate visual features to guide relational feature learning while assigning weights to valuable attributes for alignment. Meanwhile, \citeauthor{DBLP:conf/coling/LinZWSW022} \cite{DBLP:conf/kdd/ChenL00WYC22} further enhance intra-modal learning via contrastive learning and apply the KL divergence over the output distribution between joint and uni-modal embedding to reduce the modality gap. 
However, all these methods ignore the dynamical inter-modal effect for each entity. 
This is non-negligible in real world EA scenarios, as there are inevitable errors and noise in KGs (especially MMKGs) found on the Internet or professional fields, such as those containing unidentifiable images. 
Besides, intra-modal feature discrepancies (e.g., node degrees) and inter-modal source preferences (e.g., phenomenon of modality absence, imbalance or ambiguity) are also commonplace across KGs. 
In this work, we propose an effective method, MEAformer, which offers a fully dynamic meta modality hybrid strategy to address the above problems.

\vspace{-1pt}
\section{Method}
We define a MMKG as a five-tuple, i.e., $\mathcal{G}$$=$$\{\mathcal{E}, \mathcal{R}, \mathcal{A}, \mathcal{V}, \mathcal{T}\}$.
$\mathcal{E}, \mathcal{R}, \mathcal{A}$ and $\mathcal{V}$ denote the sets of entities, relations, attributes, and images, respectively. $\mathcal{T}$ $\subseteq$ $\mathcal{E} \times \mathcal{R} \times\mathcal{E}$ is the set of relation triples.
Given two MMKGs $\mathcal{G}_1$ $=$ $\{\mathcal{E}_1, \mathcal{R}_1, \mathcal{A}_1, \mathcal{V}_1, \mathcal{T}_1\}$ and $\mathcal{G}_2$ $=$ $\{\mathcal{E}_2, \mathcal{R}_2, \mathcal{A}_2, \mathcal{V}_2, \mathcal{T}_2\}$, MMEA aims to discern each entity pair ($e^1_i$, $e^2_i$), $e^1_i \in \mathcal{E}_1$, $e^2_i \in \mathcal{E}_2$ where $e^1_i$ and $e^2_i$ correspond to an identical real-world entity $e_i$.
For clarity, we omit the superscript symbol denoting the source KG of an entity in our context, except when explicitly required in statements or formulas.
Each entity is associated with multiple attributes and  $0$ or $1$ image. 
A set of pre-aligned entity pairs is provided, which are proportionally divided into a training set (seed alignments $\mathcal{S}$) and a testing set $\mathcal{S}_{te}$ based on the given seed alignment ratio ($R_{sa}$). 
Besides, we denote the available modality set as $\mathcal{M}$.

\subsection{Multi-modal Knowledge Embedding}
This section elaborates how we embed each modality $m$ of an entity into a low-dimensional vector $h^m$ in the given MMKGs.

\subsubsection{\textbf{Graph Neighborhood Structure Embedding.}} 

Let $x_i^g \in \mathbb{R}^d$ represent the  randomly initialized graph embedding of entity $e_i$ where $d$ is the predetermined hidden dimension. 
We employ a (two attention heads and two layers) Graph Attention Network (GAT) \cite{DBLP:conf/iclr/VelickovicCCRLB18} to capture the structural information of $\mathcal{G}$, equipped with a diagonal weight matrix \cite{DBLP:journals/corr/YangYHGD14a} $\bm{W}g \in \mathbb{R}^{d \times d}$ for linear transformation:
\begin{equation} \label{eq:gat}
 h_i^g = GAT(\bm{W}_g, \bm{M}_g; x_i^g)\,,
\end{equation}
where $\bm{M}_g$ denotes the graph adjacency matrix. 

\subsubsection{\textbf{Relation, Attribute, Visual, and Surface Embedding.}} 
In order to avoid the information pollution brought by mixing the representation from relations and attributes in GNN network \cite{DBLP:conf/aaai/0001CRC21}, we apply the separate fully connected layers parameterized by $\bm{W}_m \in \mathbb{R}^{d_m \times d}$ to transform those features $x^m$:
\begin{equation} \label{eq:map}
 \vspace{-1pt}
 h_i^m = FC_m(\bm{W}_m, x_i^m)\,, \,\,\, m \in \{r, a, v, s\}\,,
 \vspace{-1pt}
\end{equation}
where $r$, $a$, $v$, $s$ represent the relation, attribute, vision, and surface (a.k.a. entity name) modality, respectively. 
$x_i^m\in\mathbb{R}^{d_m}$ is the input feature of entity $e_i$ for the corresponding modality $m$.
We follow the approach of Yang et al. \cite{DBLP:conf/emnlp/YangZSLLS19} by using the bag-of-words features for relation ($x^r$) and attribute ($x^a$) representations, with their types as the minimum units.
While for the visual modality, we employ a pre-trained visual model as the encoder ($Enc_v$) to obtain the visual embeddings $x^v_i$ for each available image $v_i$ of the entity $e_i$, where the final layer output before logits serves as the image feature.
The overall model diagram is shown in Figure \ref{fig:model}, where we omit the surface encoder to increase the clarity.

\subsection{Meta Modality Hybrid}
This section presents the meta modality hybrid (MMH) module, which enables the dynamic modality fusion for entities. 

\subsubsection{\textbf{Dynamic Cross-modal Weighted (DCMW)}}
DCMW aims to dynamically generate the entity-level meta weight for each modality.
Inspired by the vanilla transformer \cite{DBLP:conf/nips/VaswaniSPUJGKP17}, we 
involve two types of sub-layers in DCMW: 
the multi-head cross-modal attention (MHCA) block and the fully connected feed-forward network (FFN).
Specifically, MHCA operates its attention function across $N_h$ parallel heads. The $i$-th head is parameterized by modally shared matrices $\bm{W}_q^{(i)}$, $\bm{W}_k^{(i)}$, $\bm{W}_v^{(i)}$ $\in \mathbb{R}^{d \times d_h}$, transforming the multi-modal input $h^m$ into modal-aware query ${Q}^{(i)}_m$, key ${K}^{(i)}_m$, and value ${V}^{(i)}_m$ in $\mathbb{R}^{d_h}$ ($d_h=d/N_h$):
\begin{equation}
{Q}^{(i)}_m, {K}^{(i)}_m, {V}^{(i)}_m =h^m \bm{W}_q^{(i)}, h^m \bm{W}_k^{(i)}, h^m \bm{W}_v^{(i)} \,.\\
\end{equation}
For the feature of modality $m$, its output is:
\begin{align}
       \operatorname{MHCA}(h^m) & =\bigoplus\nolimits_{i=1}^{N_h}\operatorname{head}i^m \cdot \bm{W}{o} \,, \\
		\operatorname{head}_{\mathrm{i}}^m & = \sum\nolimits_{j \in \mathcal{M}} \beta^{(i)}_{mj}{V}^{(i)}_j \,,
\end{align}
where $\bm{W}_{o}$ $\in \mathbb{R}^{d \times d}$ and $\bigoplus$ refers to the concatenation operation. The attention weight ($\beta_{m j}$)
 between an entity's modality $m$ and $j$  in each head  is formulated below:
\begin{equation}
    \beta_{m j}=\frac{\exp (Q_m^\top K_j / \sqrt{d_h} )}{\sum_{n \in \mathcal{M}} \exp (Q_m^\top K_n / \sqrt{d_h})}\,,
\end{equation}
where $d_h=d/N_h$.
Besides, layer normalization (LN) and residual connection (RC) are used to stabilize the training:
\begin{equation}
\hat{h}^m = LayerNorm(\operatorname{MHCA}(h^m) + h^m) \,.
\end{equation}

FFN consists of two linear transformation layers with a ReLU activation function and LN$\&$RC applied afterward as follows:
\begin{align}
\operatorname{FFN}(\hat{h}^m) & = ReLU(\hat{h}^m\bm{W}_{1} + b_{1})\bm{W}_{2} +b_{2} \,, \\
\hat{h}^m & \gets LayerNorm(\operatorname{FFN}(\hat{h}^m) + \hat{h}^m) \,,
\end{align}
where $\bm{W}_{1}$ $\in \mathbb{R}^{d \times d_{in}}$ and $\bm{W}_{2}$ $\in \mathbb{R}^{d_{in} \times d}$. We note that, in our MEAformer, the FFN is an optional component as reducing parameters can sometimes alleviate overfitting with less complex data (e.g., fewer attribute types). 

We define the output meta weight $w_m$ for each modality $m$ as:
\begin{equation}
    w_m = \frac{\exp(\sum\nolimits_{j \in \mathcal{M}} \sum\nolimits_{i=0}^{N_h}  \beta^{(i)}_{mj}/\sqrt{|\mathcal{M}| \times N_h})}{\sum\nolimits_{k \in \mathcal{M}}\exp(\sum\nolimits_{j \in \mathcal{M}} \sum\nolimits_{i=0}^{N_h}  \beta^{(i)}_{kj}\sqrt{|\mathcal{M}| \times N_h})}\,,
\end{equation}
which captures crucial inter-modal interface information, and adaptively adjusts model's cross-modal alignment preference for different modalities from each entity.

\subsubsection{\textbf{Modality Fusion}}
Let $w_m^i$ be the meta weight of entity $e_i$ for modality $m$; 
we formulate the joint embedding as:
\begin{equation} \label{eq:cat}
    {h}^{\mu}_i=\bigoplus\nolimits_{m \in \mathcal{M}}[w^i_m{h}_i^m]\,, \,\,\,{{h}}^{\xi}_i=\bigoplus\nolimits_{m \in \mathcal{M}}[w^i_m\hat{h}_i^m]\,,
\end{equation}
where ${{h}}^{\mu}_i$ and ${{h}}^{\xi}_i$ are defined as the {\emph{early and late fusion embedding}}, respectively. As a common paradigm for modality fusion in previous EA works \cite{DBLP:conf/coling/LinZWSW022,DBLP:conf/aaai/0001CRC21,DBLP:conf/kdd/ChenL00WYC22}, concatenation operation could be employed to prevent obscuring and over-smoothing modal diversities.
We select ${{h}}^{\mu}_i$ as the final entity representation for evaluations according to our experiments, and we speculate that the modality specificity in late fusion embedding is consistently attenuated by the Transformer Layer \cite{DBLP:journals/corr/abs-2103-11886} which diminishes the distinction between entities.

\subsection{Modal-adaptive Contrastive Learning.}
In this section, we introduce modal-adaptive contrastive learning to sufficiently mine the information contained in each modality with the limited seed alignments $\mathcal{S}$.

Specifically, we corrupt the seed alignments $\mathcal{S}$ for negative alignments, following the 1-to-1 alignment assumption used in EA research \cite{DBLP:conf/ijcai/SunHZQ18}.
For each entity pair ($e_i^1$,$e_i^2$) in $\mathcal{S}$,
we define $\mathcal{N}^{ng}_i$ $=$ $\{e^1_j|\forall e^1_j \in \mathcal{E}_1, j \neq i\}$ $\cup$ $\{e^2_j|\forall e^2_j \in \mathcal{E}_2, j \neq i\}$ as its negative entity set.
 Additionally, we employ the in-batch negative sampling strategy \cite{DBLP:conf/icml/ChenK0H20} to limit the sampling scope of $\mathcal{N}^{ng}_i$ within the mini-batch for efficiency.
The alignment probability 
is defined as:
\begin{equation}\label{eq:macl}
    p_m(e^1_i, e^2_i) = \frac{\gamma_m(e^1_i, e^2_i)}{\gamma_m(e^1_i, e^2_i) + \sum\nolimits_{e_j \in \mathcal{N}^{ng}_i}\gamma_m(e^1_i, e_j)} \, ,
\end{equation}
where $\gamma_m(e_i, e_j)$ $=$ $\exp({h^{m\top}_{i}}{h^m_{j}}/\tau)$ and $\tau$ is the temperature hyper-parameter.
Considering the alignment direction for entity pairs reflected in \eqref{eq:macl}, we define the bi-directional alignment objective for each modality ($m$) as:
\begin{equation} \label{eq:loss}
    \mathcal{L}_m = -log(p_m(e^1_i, e^2_i)+p_m(e^2_i, e^1_i))/2 \,.
\end{equation}
Furthermore, we introduce the intra-modal contrastive loss $\mathcal{L}_{ICL}$ which encourages the model to explicitly align the multi-modal feature $h^m$ \cite{DBLP:conf/coling/LinZWSW022}.
Meanwhile, we involve a late intra-modal contrastive loss $\mathcal{L}_{LICL}$ to enable the mutual information complementarity among modalities via cross-modal attentive knowledge transfer:
\begin{equation}
    \mathcal{L}_{ICL} = \sum\nolimits_{m \in \mathcal{M}}\mathcal{L}_m,\,\,
    \mathcal{L}_{LICL} = \sum\nolimits_{m \in \mathcal{M}}\widehat{\mathcal{L}}_m,
\end{equation}
where $\widehat{\mathcal{L}}_m$ is $\mathcal{L}_m$'s variant with $\widehat{\gamma}_m(e_i, e_j)$$=$$\exp(\hat{h}^{m\top}_{i}{\hat{h}^m_{j}}/\tau)$.

Finally, we train the network to minimize the overall loss:
\begin{equation}
    \mathcal{L} = \mathcal{L}_{\mu} + \mathcal{L}_{ICL} + \mathcal{L}_{LICL},
\end{equation}
where$ \mathcal{L}_{\mu}$ is based on the joint early fusion embedding $h^{\mu}$, following Equation \eqref{eq:loss} with $m=\mu$.

\subsubsection{\textbf{modal-aware Hard Entity Replay.}}
To further improve the model’s performance on those hard  entity pairs (i.e., those with high likelihood of alignment but possessing ambiguous details),
we propose a modal-aware  entity replay (MERP) policy.
Concretely, it requires the model to sample not only in-batch negatives but also to actively search for out-of-batch hard negatives.  
To accelerate this process, a hard negative matrix $\bm{M}_{neg} \in \mathbb{R}^{(|\mathcal{E}_1|+|\mathcal{E}_2|) \times 2}$ is adopted, which stores and updates the nearest non-aligned target for each entity in the vector space after each training step. 
Given a mini-batch of entities, those existing hard negative entities can be quickly retrieved by indexing $\bm{M}_{neg}$, thereby expanding the original negative set.  
In order to mitigate the noise introduced by independent modality divergence (i.e, similar entity pairs may exhibit inconsistent similarity in a certain modality), we apply MERP solely to the confluent objective $\mathcal{L}_{\mu}$, which is based on early joint embedding.

\vspace{-1pt}
\begin{table*}[ht]
    \centering
	\tabcolsep=0.3cm
    \renewcommand\arraystretch{1.0}
        \caption{\textbf{Non-iterative} results  without (w/o) and with (w/) surface forms (SF) on three \textbf{bilingual} datasets, 
    where $\dagger$ denotes that the PLMs were applied for surface/attribute embedding generation.  
    ``Para.'' refers to the number of learnable parameters.
    ``Unsup.'' refers to the unsupervised learning depending on entities’ surface (S) or visual (V) information. 
    The best results in baselines are marked with \underline{underline}, and we highlight our results with \textbf{bold} when we achieve new SOTA.
    The symbol $\ast$ denotes our reproduced results on the same experiment setting.
    }
    \label{tab:overall-no-iter}
    \vspace{-2pt}
    \resizebox{0.73\linewidth}{!}{
    \begin{tabular}{@{}l|l|c|ccc|ccc|ccc}
        \toprule
        & \multirow{2}*{\makebox[2.6cm][c]{Models}} & \multirow{2}*{Para.} & \multicolumn{3}{c|}{DBP15K$_{ZH-EN}$} & \multicolumn{3}{c|}{DBP15K$_{JA-EN}$} & \multicolumn{3}{c}{DBP15K$_{FR-EN}$} \\
        & & & {\scriptsize H@1} & {\scriptsize H@10} & {\scriptsize MRR} & {\scriptsize H@1} & {\scriptsize H@10} & {\scriptsize MRR} & {\scriptsize H@1} & {\scriptsize H@10} & {\scriptsize MRR} \\
        \midrule
        \parbox[t]{2mm}{\multirow{7}{*}{\rotatebox[origin=c]{90}{w/o SF}}} 
        & MUGNN {\footnotesize \cite{DBLP:conf/acl/CaoLLLLC19}} & - &
        .494 & .844 & .611 &  .501 & .857 & .621 & .495 & .870 & .621 \\
        & AliNet {\footnotesize {\cite{DBLP:conf/aaai/SunW0CDZQ20}}} & - &
        .539 & .826 & .628 & .549 & .831 & .645 & .552 & .852 & .657 \\
        & EVA* {\footnotesize \cite{DBLP:conf/aaai/0001CRC21}} & 13.3M &
        {.680} & {.910} & {.762} & {.673} & {.908} & {.757} & {.683} & \underline{.923} & {.767} \\
        & MSNEA* {\footnotesize {\cite{DBLP:conf/kdd/ChenL00WYC22}}} & 14.1M & .601 & .830 & .684 & .535 & .775 & .617 & .543 & .801 & .630 \\
        & MCLEA* {\footnotesize {\cite{DBLP:conf/coling/LinZWSW022}}} & 13.2M &
        \underline{.715} & \underline{.923} & \underline{.788} & \underline{.715} & \underline{.909} & \underline{.785} & \underline{.711} & {.909} & \underline{.782} \\
        & \CC\textbf{MEAformer} {\footnotesize (Ours)} & \CC13.7M &
        \CC\textbf{.771} & \CC\textbf{.951} & \CC\textbf{.835} & \CC\textbf{.764} & \CC\textbf{.959} & \CC\textbf{.834} & \CC\textbf{.770} & \CC\textbf{.961} & \CC\textbf{.841} \\
		& \CK~~~~~w/ MERP  & \CK13.7M & \CK\textbf{.772} & \CK\textbf{.951} & \CK\textbf{.835} & \CK\textbf{.769} & \CK\textbf{.961} & \CK\textbf{.840} & \CK\textbf{.771} & \CK\textbf{.962} & \CK\textbf{.841} \\
        \midrule
        \parbox[t]{2mm}{\multirow{10}{*}{\rotatebox[origin=c]{90}{w/ SF}}} 
        & AttrGNN$\dagger$ {\footnotesize \cite{DBLP:conf/emnlp/LiuCPLC20}} & - &
        .777 & .920 & .829 & .763 & .909 & .816 & .942 & .987 & .959 \\
		& RNM {\footnotesize \cite{DBLP:conf/aaai/00020WD21}} & - & .840 & .919 & .870 & .872 & .944 & .899 & .938 & .981 & .954  \\
		& CLEM$\dagger$ {\footnotesize \cite{wu2022leveraging}} & - & .854 & .935 & .879 & .885 & .958 & .904 & .936 & .977 & .952 \\
  		& RPR-RHGT {\footnotesize \cite{DBLP:conf/ijcai/CaiMZJ22}} & - & .693 & - & .754 & .886 & - & .912 & .889 & - & .919 \\
        & ERMC$\dagger$ {\footnotesize \cite{DBLP:conf/cikm/YangWZQWHH21}} & - &
        .903 & .946 & .899 & .942 & .944 & .925 & .962 & .982 & .973 \\
        & EVA* {\footnotesize \cite{DBLP:conf/aaai/0001CRC21}} & 13.8M & \underline{.929} & \underline{.986} & \underline{.951} & \underline{.964} & \underline{.997} & \underline{.976} & \underline{.990} & \underline{.999} & \underline{.994} \\
        & MSNEA* {\footnotesize {\cite{DBLP:conf/kdd/ChenL00WYC22}}} & 14.7M & .887 & .961 & .913 & .938 & .983 & .955 & .969 & .997 & .980 \\
        & MCLEA* {\footnotesize {
        \cite{DBLP:conf/coling/LinZWSW022}}} & 13.7M &
        {.926} & {.983} & {.946} & {.961} & {.994} & {.973} & {.987} & \underline{.999} & {.992} \\   
                & \CC\textbf{MEAformer} {\footnotesize (Ours)} & \CC14.2M &
        \CC\textbf{.948} & \CC\textbf{.993} & \CC\textbf{.965} & \CC\textbf{.977} & \CC\textbf{.999} & \CC\textbf{.986} & \CC\textbf{.991} & \CC\textbf{1.00} & \CC\textbf{.995} \\
		& \CK~~~~~w/ MERP  & \CK14.2M & \CK\textbf{.949} & \CK\textbf{.993} & \CK\textbf{.965} & \CK\textbf{.978} & \CK\textbf{.999} & \CK\textbf{.986} & \CK\textbf{.991} & \CK\textbf{1.00} & \CK\textbf{.995} \\
		\midrule
		\parbox[t]{2mm}{\multirow{4}{*}{\rotatebox[origin=c]{90}{(S) Unsup.}}} 
		& EVA* {\footnotesize \cite{DBLP:conf/aaai/0001CRC21}} & 13.8M & \underline{.883} & \underline{.967} & \underline{.913} & .930 & \underline{.985} & \underline{.951} & \underline{.968} & \underline{.995} & \underline{.978} \\
        & MSNEA* {\footnotesize {\cite{DBLP:conf/kdd/ChenL00WYC22}}} & 14.7M & .858 & .935 & .886 & .921 & .973 & .939 & .953 & .990 & .967 \\
        & MCLEA* {\footnotesize {\cite{DBLP:conf/coling/LinZWSW022}}} & 13.7M &
        {.879} & {.963} & {.909} & \underline{.931} & {.983} & \underline{.951} & {.959} & {.993} & {.972} \\
        & \CC\textbf{MEAformer} {\footnotesize (Ours)} & \CC14.2M &
        \CC\textbf{.917} & \CC\textbf{.980} & \CC\textbf{.941} & \CC\textbf{.958} & \CC\textbf{.992} & \CC\textbf{.972} & \CC\textbf{.973} & \CC\textbf{.998} & \CC\textbf{.982} \\
		\midrule
		\parbox[t]{2mm}{\multirow{4}{*}{\rotatebox[origin=c]{90}{(V) Unsup.}}}  
		& EVA* {\footnotesize \cite{DBLP:conf/aaai/0001CRC21}} & 13.8M & \underline{.891} & \underline{.961} & \underline{.917} & \underline{.941} & \underline{.986} & \underline{.958} & \underline{.970} & \underline{996} & \underline{.982} \\
        & MSNEA* {\footnotesize {\cite{DBLP:conf/kdd/ChenL00WYC22}}} & 14.7M & .859 & .936 & .887 & .921 & .970 & .939 & .954 & .989 & .968 \\
        & MCLEA* {\footnotesize {\cite{DBLP:conf/coling/LinZWSW022}}} & 13.7M &
        {.860} & {.950} & {.893} & {.914} & {.975} & {.938} & {.953} & {.990} & {.967} \\
        & \CC\textbf{MEAformer} {\footnotesize (Ours)} & \CC14.2M &
        \CC\textbf{.909} & \CC\textbf{.974} & \CC\textbf{.933} & \CC\textbf{.950} & \CC\textbf{.990} & \CC\textbf{.965} & \CC\textbf{.972} & \CC\textbf{.997} & \CC\textbf{.983} \\
        \bottomrule
    \end{tabular}
    \vspace{-4pt}
    }
\end{table*}

\begin{table*}[ht]
    \centering
	\tabcolsep=0.3cm
    \renewcommand\arraystretch{1.0}
    
    \caption{\textbf{Iterative} results on three \textbf{bilingual} datasets.}
    \vspace{-3pt}
    \resizebox{0.73\linewidth}{!}{
    \begin{tabular}{@{}l|l|c|ccc|ccc|ccc}
        \toprule
        & \multirow{2}*{\makebox[2.6cm][c]{Models}} & \multirow{2}*{Para.} & \multicolumn{3}{c|}{DBP15K$_{ZH-EN}$} & \multicolumn{3}{c|}{DBP15K$_{JA-EN}$} & \multicolumn{3}{c}{DBP15K$_{FR-EN}$} \\
        & & & {\scriptsize H@1} & {\scriptsize H@10} & {\scriptsize MRR} & {\scriptsize H@1} & {\scriptsize H@10} & {\scriptsize MRR} & {\scriptsize H@1} & {\scriptsize H@10} & {\scriptsize MRR} \\
        \midrule
        \parbox[t]{2mm}{\multirow{6}{*}{\rotatebox[origin=c]{90}{w/o SF}}} 
        & BootEA {\footnotesize \cite{DBLP:conf/ijcai/SunHZQ18}} & - & 
        .629 & .847 & .703 & .622 & .854 & .701 & .653 & .874 & .731 \\
        & NAEA {\footnotesize {\cite{DBLP:conf/ijcai/ZhuZ0TG19}}} & - &
        .650 & .867 & .720 & .641 & .873 & .718 & .673 & .894 & .752 \\
        & EVA* {\footnotesize \cite{DBLP:conf/aaai/0001CRC21}} & 13.3M &
        {.746} & {.910} & {.807} & {.741} & {.918} & {.805} & {.767} & {.939} & {.831} \\
        & MSNEA* {\footnotesize {\cite{DBLP:conf/kdd/ChenL00WYC22}}} & 14.1M & .643 & .865 & .719 & .572 & .832 & .660 & .584 & .841 & .671 \\
        & MCLEA* {\footnotesize {\cite{DBLP:conf/coling/LinZWSW022}}} & 13.2M &
        \underline{.811} & \underline{.954} & \underline{.865} & \underline{.806} & \underline{.953} & \underline{.861} & \underline{.811} & \underline{.954} & \underline{.865} \\
        & \CC\textbf{MEAformer} {(Ours)} & \CC13.7M &
        \CC\textbf{.847} & \CC\textbf{.970} & \CC\textbf{.892} & \CC\textbf{.842} & \CC\textbf{.974} & \CC\textbf{.892} & \CC\textbf{.845} & \CC\textbf{.976} & \CC\textbf{.894} \\
        \midrule
        \parbox[t]{2mm}{\multirow{4}{*}{\rotatebox[origin=c]{90}{w/ SF}}} 
        & EVA* {\footnotesize \cite{DBLP:conf/aaai/0001CRC21}} & 13.8M & .956 & .993 & .969 & .979 & .998 & .987 & \underline{.995} & .999 & \underline{.997} \\
        & MSNEA* {\footnotesize {\cite{DBLP:conf/kdd/ChenL00WYC22}}} & 14.7M & .896 & .969 & .922 & .942 & .986 & .958 & .971 & .998 & .982 \\
        & MCLEA* {\footnotesize {\cite{DBLP:conf/coling/LinZWSW022}}} & 13.7M &
        \underline{.964} & \underline{.996} & \underline{.977} & \underline{.986} & \underline{.999} & \underline{.992} & \underline{.995} & \underline{1.00} & \underline{.997} \\
        & \CC\textbf{MEAformer} {\footnotesize (Ours)} & \CC14.2M &
        \CC\textbf{.973} & \CC\textbf{.998} & \CC\textbf{.983} & \CC\textbf{.991} & \CC\textbf{1.00} & \CC\textbf{.995} & \CC\textbf{.996} & \CC\textbf{1.00} & \CC\textbf{.998} \\
		\midrule
		\parbox[t]{2mm}{\multirow{5}{*}{\rotatebox[origin=c]{90}{(S) Unsup.}}} 
        & EASY$\dagger$ {\footnotesize {\cite{DBLP:conf/sigir/GeLCZG21}}} & - &
        {.898} & {.979} & {.930} & {.943} & {.990} & {.960} & {.980} & {.998} & {.990} \\
		& EVA* {\footnotesize \cite{DBLP:conf/aaai/0001CRC21}} & 13.8M & .937 & .991 & .957 & .974 & .998 & .983 & \underline{.992} & \underline{1.00} & \underline{.996} \\
        & MSNEA* {\footnotesize {\cite{DBLP:conf/kdd/ChenL00WYC22}}} & 14.7M & .870 & .946 & .897 & .933 & .980 & .950 & .961 & .992 & .973 \\
        & MCLEA* {\footnotesize {\cite{DBLP:conf/coling/LinZWSW022}}} & 13.7M &
        \underline{.947} & \underline{.995} & \underline{.966} & \underline{.977} & \underline{.999} & \underline{.986} & {.990} & \underline{1.00} & {.994} \\
        & \CC\textbf{MEAformer} {\footnotesize (Ours)} & \CC14.2M &
        \CC\textbf{.962} & \CC\textbf{.998} & \CC\textbf{.976} & \CC\textbf{.987} & \CC\textbf{.999} & \CC\textbf{.992} & \CC\textbf{.993} & \CC\textbf{1.00} & \CC\textbf{.996} \\
		\midrule
		\parbox[t]{2mm}{\multirow{4}{*}{\rotatebox[origin=c]{90}{(V) Unsup.}}} 
		& EVA* {\footnotesize \cite{DBLP:conf/aaai/0001CRC21}} & 13.8M & \underline{.948} & .992 & \underline{.964} & \underline{.977} & .997 & \underline{.985} & \underline{.990} & \underline{1.00} & \underline{.995} \\
        & MSNEA* {\footnotesize {\cite{DBLP:conf/kdd/ChenL00WYC22}}} & 14.7M & .871 & .943 & .898 & .927 & .976 & .945 & .959 & .993 & .979 \\
        & MCLEA* {\footnotesize {\cite{DBLP:conf/coling/LinZWSW022}}} & 13.7M &
        {.942} & \underline{.994} & {.963} & {.974} & \underline{.999} & {.984} & {.986} & \underline{1.00} & {.992} \\
        & \CC\textbf{MEAformer} {\footnotesize (Ours)} & \CC14.2M &
        \CC\textbf{.964} & \CC\textbf{.998} & \CC\textbf{.975} & \CC\textbf{.985} & \CC\textbf{1.00} & \CC\textbf{.991} & \CC\textbf{.992} & \CC\textbf{1.00} & \CC\textbf{.996} \\
        \bottomrule
    \end{tabular}
    }
    
    \label{tab:overall-iter}
    \vspace{-3pt}
\end{table*}

\begin{table}[ht]
    \centering
    \renewcommand\arraystretch{1.0}
    \caption{\textbf{Non-iterative} results on two \textbf{monolingual} datasets compared with MMEA methods where \textit{X}\% represents the percentage of reference entity alignments used for training. 
    }
    \label{tab:overall-non-iter-FBDBYG}
    \vspace{-1pt}
    \resizebox{0.80\linewidth}{!}{
    \begin{tabular}{@{}l|l|ccc|ccc}
        \toprule
        & \multirow{2}*{\makebox[1.8cm][c]{Models}} & \multicolumn{3}{c|}{FB15K-DB15K} & \multicolumn{3}{c}{FB15K-YAGO15K} \\
        & & {\scriptsize H@1} & {\scriptsize H@10} & {\scriptsize MRR} & {\scriptsize H@1} & {\scriptsize H@10} & {\scriptsize MRR} \\
        \midrule
        \parbox[t]{1.2mm}{\multirow{6}{*}{\rotatebox[origin=c]{90}{20\%}}} 
        & MMEA & {.265} & {.541} & {.357} & {.234} & {.480} & {.317} \\
        & EVA$\ast$ & .199 & .448 & .283 & .153 & .361 & .224 \\
        & MSNEA$\ast$ & .114 & .296 & .175 & .103 & .249 & .153 \\
        & MCLEA$\ast$ & \underline{.295} & \underline{.582} & \underline{.393} & \underline{.254} & \underline{.484} & \underline{.332} \\
        & \CC\textbf{MEAformer} & \CC\textbf{.417} & \CC\textbf{.715} & \CC\textbf{.518} & \CC\textbf{.327} & \CC\textbf{.595} & \CC\textbf{.417} \\
        & \CK\textbf{~~~~~w/ MERP} & \CK\textbf{.434} & \CK\textbf{.728} & \CK\textbf{.534} & \CK\textbf{.325} & \CK\textbf{.598} & \CK\textbf{.416} \\
        \midrule
        \parbox[t]{1.2mm}{\multirow{6}{*}{\rotatebox[origin=c]{90}{50\%}}} 
        & MMEA & .417 & {.703} & .512 & {.403} & {.645} & {.486} \\
        & EVA$\ast$ & .334 & .589 & .422 & .311 & .534 & .388 \\
        & MSNEA$\ast$ & .288 & .590 & .388 & .320 & .589 & .413 \\
        & MCLEA$\ast$ & \underline{.555} & \underline{.784} & \underline{.637} & \underline{.501} & \underline{.705} & \underline{.574} \\
        & \CC\textbf{MEAformer} & \CC\textbf{.619} & \CC\textbf{.843} & \CC\textbf{.698} & \CC\textbf{.560} & \CC\textbf{.778} & \CC\textbf{.639} \\
        & \CK\textbf{~~~~~w/ MERP} & \CK\textbf{.625} & \CK\textbf{.847} & \CK\textbf{.704} & \CK\textbf{.560} & \CK\textbf{.780} & \CK\textbf{.640} \\
        \midrule
        \parbox[t]{1.2mm}{\multirow{6}{*}{\rotatebox[origin=c]{90}{80\%}}} 
        & MMEA & .590 & {.869} & .685 & {.598} & \underline{.839} & {.682} \\
        & EVA$\ast$ & .484 & .696 & .563 & .491 & .692 & .565 \\
        & MSNEA$\ast$ & .518 & .779 & .613 & .531 & .778 & .620 \\
        & MCLEA$\ast$ & \underline{.735} & \underline{.890} & \underline{.790} & \underline{.667} & {.824} & \underline{.722} \\
        & \CC\textbf{MEAformer} & \CC\textbf{.765} & \CC\textbf{.916} & \CC\textbf{.820} & \CC\textbf{.703} & \CC\textbf{.873} & \CC\textbf{.766} \\
        & \CK\textbf{~~~~~w/ MERP} & \CK\textbf{.773} & \CK\textbf{.918} & \CK\textbf{.825} & \CK\textbf{.705} & \CK\textbf{.874} & \CK\textbf{.768} \\
        \bottomrule
    \end{tabular}
    }
    \vspace{-8pt}
\end{table}

\begin{table}[ht]
    \centering
    \renewcommand\arraystretch{1.0}
    \caption{\textbf{Iterative} results on two \textbf{monolingual} datasets.
    }
    \label{tab:overall-iter-FBDBYG}
    \vspace{-1pt}
    \resizebox{0.80\linewidth}{!}{
    \begin{tabular}{@{}l|l|ccc|ccc}
        \toprule
        & \multirow{2}*{\makebox[1.8cm][c]{Models}} & \multicolumn{3}{c|}{FB15K-DB15K} & \multicolumn{3}{c}{FB15K-YAGO15K} \\
        & & {\scriptsize H@1} & {\scriptsize H@10} & {\scriptsize MRR} & {\scriptsize H@1} & {\scriptsize H@10} & {\scriptsize MRR} \\
        \midrule
        \parbox[t]{1.2mm}{\multirow{4}{*}{\rotatebox[origin=c]{90}{20\%}}} 
        & EVA$\ast$ & .231 & .488 & .318 & .188 & .403 & .260 \\
        & MSNEA$\ast$ & .149 & .392 & .232 & .138 & .346 & .21 \\
        & MCLEA$\ast$ & \underline{.395} & \underline{.656} & \underline{.487} & \underline{.322} & \underline{.546} & \underline{.400} \\
        & \CC\textbf{MEAformer} & \CC\textbf{.578} & \CC\textbf{.812} & \CC\textbf{.661} & \CC\textbf{.444} & \CC\textbf{.692} & \CC\textbf{.529} \\
        \midrule
        \parbox[t]{1.2mm}{\multirow{4}{*}{\rotatebox[origin=c]{90}{50\%}}} 
        & EVA$\ast$ & .364 & .606 & .449 & .325 & .560 & .404 \\
        & MSNEA$\ast$ & .358 & .656 & .459 & .376 & .646 & .472 \\
        & MCLEA$\ast$ & \underline{.620} & \underline{.832} & \underline{.696} & \underline{.563} & \underline{.751} & \underline{.631} \\
        & \CC\textbf{MEAformer} & \CC\textbf{.690} & \CC\textbf{.871} & \CC\textbf{.755} & \CC\textbf{.612} & \CC\textbf{.808} & \CC\textbf{.682} \\
        \midrule
        \parbox[t]{1.2mm}{\multirow{4}{*}{\rotatebox[origin=c]{90}{80\%}}} 
        & EVA$\ast$ & .491 & .711 & .573 & .493 & .695 & .572 \\
        & MSNEA$\ast$ & .565 & .810 & .651 & .593 & .806 & .668 \\
        & MCLEA$\ast$ & \underline{.741} & \underline{.900} & \underline{.802} & \underline{.681} & \underline{.837} & \underline{.737} \\
        & \CC\textbf{MEAformer} & \CC\textbf{.784} & \CC\textbf{.921} & \CC\textbf{.834} & \CC\textbf{.724} & \CC\textbf{.880} & \CC\textbf{.783} \\
        \bottomrule
    \end{tabular}
    }
    \vspace{-8pt}
\end{table}
\section{Experiment}
\vspace{-1pt}
All ablation, efficiency analyses, and case studies are conducted  in the typical supervised scenario with the hidden dimension $d$ set to 300 unless otherwise specified.
Please refer to the appendix for details about the different experimental scenarios and metrics.

\subsection{Experiment Setup}
\subsubsection{\textbf{Datasets.}}
We consider two types of datasets.
\textbf{\textit{(\rmnum{1})}} {\emph{Bilingual}}: DBP15K \cite{DBLP:conf/semweb/SunHL17} contains three  datasets built from the multilingual versions of DBpedia, including DBP15K$_{ZH\text{-}EN}$, DBP15K$_{JA\text{-}EN}$ and DBP15K$_{FR\text{-}EN}$. Each of them contains about $400$K triples and $15$K pre-aligned entity pairs with 30\% of them as the seed alignments ($R_{sa}=0.3$). We adopt their multi-model variant \cite{DBLP:conf/aaai/0001CRC21} with images attached to the entities.
\textbf{\textit{(\rmnum{2})}} {\emph{Monolingual}}: We select FB15K-DB15K (FBDB15K) and FB15K-YAGO15K (FBYG15K) from MMKG \cite{DBLP:conf/esws/LiuLGNOR19} with three data splits: $R_{sa} \in \{0.2, 0.5, 0.8\}$.
%
For the entities without associated images, 
random vectors are generated as their visual feature using a normal distribution parameterised by the mean and standard deviation of other available images
\cite{DBLP:conf/aaai/0001CRC21}. 

\subsubsection{\textbf{Iterative Training.}}
Following \citeauthor{DBLP:conf/coling/LinZWSW022} \cite{DBLP:conf/coling/LinZWSW022}, we adopt a probation technique for iterative training.
Concretely,  every $K_e$ (where $K_e = 5$) epochs, we propose cross-KG entity pairs that are mutual nearest neighbors in the vector space and add them to a candidate list $\mathcal{N}^{cd}$. 
Furthermore, an entity pair in $\mathcal{N}^{cd}$ will be added into the training set if it remains a mutual nearest neighbour for $K_s$ ($=$ $10$) consecutive rounds.

\subsubsection{\textbf{Unsupervised Training.}}
Unsupervised MMEA was first introduced by \citeauthor{DBLP:conf/aaai/0001CRC21} \cite{DBLP:conf/aaai/0001CRC21}, utilizing the visual similarity of entities to construct a pseudo seed dictionary ($\mathcal{S}_{dic}$) to eliminate the reliance on gold labels.
Note that $\mathcal{S}_{dic}$ is initially an empty set with a predefined maximum size ($N_{dic}$).  
The algorithm then populates the dictionary $\mathcal{S}_{dic}$ up to the length $N_{dic}$ 
based on the sorted similarity of pre-extracted modality features for each entity. Surface and visual modality information serve as entity reference to obtain unsupervised results.

\subsubsection{\textbf{Baselines.}}
$13$ prominent EA algorithms proposed in recent years are selected as our basic baselines. For a clear comparison, we categorize them into four groups based on their iterability and usage of surface format information.  
Moreover,
we further collect 5 MMEA methods as strong baselines, including CLEM \cite{wu2022leveraging}, MMEA \cite{DBLP:conf/ksem/ChenLWXWC20}, EVA \cite{DBLP:conf/aaai/0001CRC21}, MSNEA \cite{DBLP:conf/kdd/ChenL00WYC22}, and MCLEA \cite{DBLP:conf/coling/LinZWSW022}.
We reproduce EVA, MSNEA, and MCLEA with their original pipelines  unchanged.

\subsubsection{\textbf{Implementation Details.}}
Due to the varying experimental datasets and environments across the aforementioned works, we reproduce them with the following settings to ensure fairness and consistency with our model:
\textbf{\textit{(\rmnum{1})}} All networks have a hidden layer dimension of 300.  
The total epochs are set to 500 with an optional iterative training strategy applied for another 500 epochs, following \cite{DBLP:conf/coling/LinZWSW022}.
Training strategies, including cosine warm-up schedule ($15\%$ steps for LR warm up), early stopping, and gradient accumulation, are adopted. The AdamW optimizer ($\beta_1=0.9$, $\beta_2=0.999$) is used with a fixed batch size of 3500. 
\textbf{\textit{(\rmnum{2})}} To demonstrate the model's stability, following \cite{DBLP:conf/ksem/ChenLWXWC20,DBLP:conf/coling/LinZWSW022}, the vision encoders $Enc_{v}$ are set to ResNet-152 \cite{DBLP:conf/cvpr/HeZRS16} on DBP15K following EVA/MCLEA, where the vision feature dimension $d_v$ is $2048$, and set to VGG-16 \cite{DBLP:journals/corr/SimonyanZ14a} on FBDB15K/FBYG15K with $d_v=4096$. 
\textbf{\textit{(\rmnum{3})}} Following \cite{DBLP:conf/emnlp/YangZSLLS19}, the Bag-of-Words (BoW) is selected for encoding relations ($x^r$) and attributes ($x^a$)  as fixed-length (e.g., $d_r=1000$) vectors. 
We use the pre-trained 300-d GloVe vectors together with the character bigrams for surface  representation after applying machine translations for entity names, as described in \cite{DBLP:conf/emnlp/MaoWWL21}. 
\textbf{\textit{(\rmnum{4})}} An alignment editing method is employed to reduce the error accumulation \cite{DBLP:conf/ijcai/SunHZQ18}.

Specifically, for MEAformer, the intermediate dimension $d_{in}$ in FFN is set to $400$.
$\tau$ is set to 0.1, and the head number $N_h$ in MHCA is set to $1$.
We eliminate the attribute values for input consistency, and extend MSNEA with iterative training capability.

We claim that the surface information will only be involved in bilingual datasets (DBP15K) following \citeauthor{DBLP:conf/coling/LinZWSW022}, and we exclude those extra entity descriptions referred in minority EA works \cite{DBLP:conf/ijcai/Tang0C00L20}.
Experiments are conducted on RTX 3090Ti GPUs without parallel.

\vspace{-4pt}
\subsection{Overall Results} \label{sec:overall}
The results of the bilingual datasets are presented in Table \ref{tab:overall-no-iter} (non-iterative) and Table \ref{tab:overall-iter} (iterative), while the results for monolingual datasets are shown in Table \ref{tab:overall-non-iter-FBDBYG} (non-iterative) and Table \ref{tab:overall-iter-FBDBYG} (iterative).
It is  evident that our model outperforms the baselines across all datasets under all metrics. 
Particularly, MEAformer surpasses those methods in typical \textbf{supervised} scenario by a large margin on Hits$@$1 of DBP15K ($5$ $\sim$ $6\%$) and FBDB15K/FBYG15K ($3$ $\sim$ $11\%$). Moreover, it consistently gains improvements on other settings compared to the SOTA methods, while maintaining a similar number of learnable parameters (approximately 14M).

It is reasonable that all models have their performance improved when incorporating the \textbf{surface} modality, as textual information inherently serves as a strong supervisory signal in all alignment fields \cite{DBLP:conf/acl/XuWYFSWY19,DBLP:conf/wsdm/MaoWXLW20}.   Nevertheless, our model still exceeds those high-performing  baselines and increases the current SOTA Hits$@$1 scores from ($.929/.964/.990$) to ($.948/.977/.991$) on $ZH$-$EN$/$JA$-$EN$/$FR$-$EN$ datasets within the DBP15K, respectively.  

As an important real world application, \textbf{unsupervised} MMEA effectively eliminates the need for manual labeling and annotation costs by generating pseudo-labeled seed alignments based on the similarity of features extracted by pre-trained models.
Concretely, we follow \citeauthor{DBLP:conf/coling/LinZWSW022} \cite{DBLP:conf/coling/LinZWSW022} to take entities' surface (S) or visual (V) features as the unsupervised references. 
The results are illustrated in Table \ref{tab:overall-no-iter} and \ref{tab:overall-iter} with ``Unsup.'' as the indication, where our MEAformer consistently achieves the SOTA results.

We note that the \textbf{MERP} policy could further increase model's performance across all datasets but it is not involved in our standard MEAformer, since we aim to keep it separate from the model and establish it as a generic technique for (MM)EA methods.
Furthermore, we abandon MERP when training model with iterative or unsupervised strategies, since
the generated pseudo seed alignments may not be entirely accurate and could mislead the model, exacerbating the error cascade issue throughout the replay stage.



\subsubsection{\textbf{Efficiency Analysis.}}
\begin{figure}[!htbp]
\vspace{-1pt}
  \centering
  \includegraphics[trim=10 0 10 0, width=0.95\linewidth]{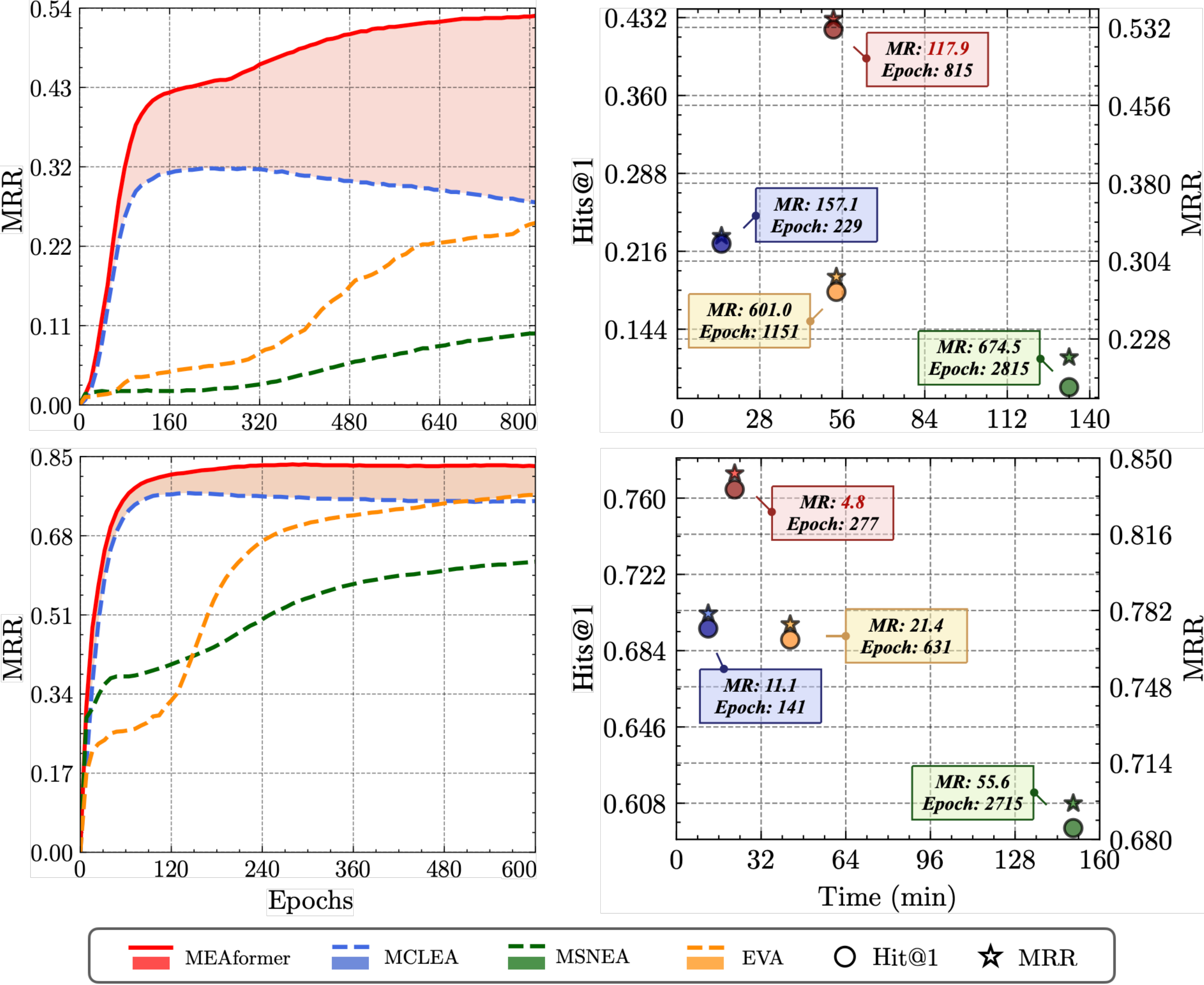}
  \vspace{-2pt}
  \caption{\textbf{Efficiency Analysis.} Performance ${vs.}$ training epoches (left), and performance ${vs.}$ training time (right), with 20$\%$ of the pre-aligned EA pairs of FBDB15K (up) and DBP15K$_{ZH-EN}$  (down).}
 \vspace{-4pt}
  \label{fig:effic}
\end{figure}

To gain further insights into our model, we examine the efficiency behaviors of four MMEA algorithms on two datasets with identical $3000$ epochs and an early stopping strategy.
As shown in Figure \ref{fig:effic}, MEAformer consistently outperforms others throughout the training progress (left),
and it also excels in the trade-off between convergence time and performance (right).
We attribute MSNEA's subpar behavior to 
the constraints imposed by translation-based models' geometric models, which  limit their ability to capture complex structural relationships among entities for EA.
Please see the appendix for details.

\subsection{Ablation Studies} \label{sec:ablation}
\subsubsection{\textbf{Low Resource.}}
\begin{figure}[!htbp]
\vspace{-1pt}
  \centering
  \includegraphics[trim=18 0 0 0, width=0.94\linewidth]{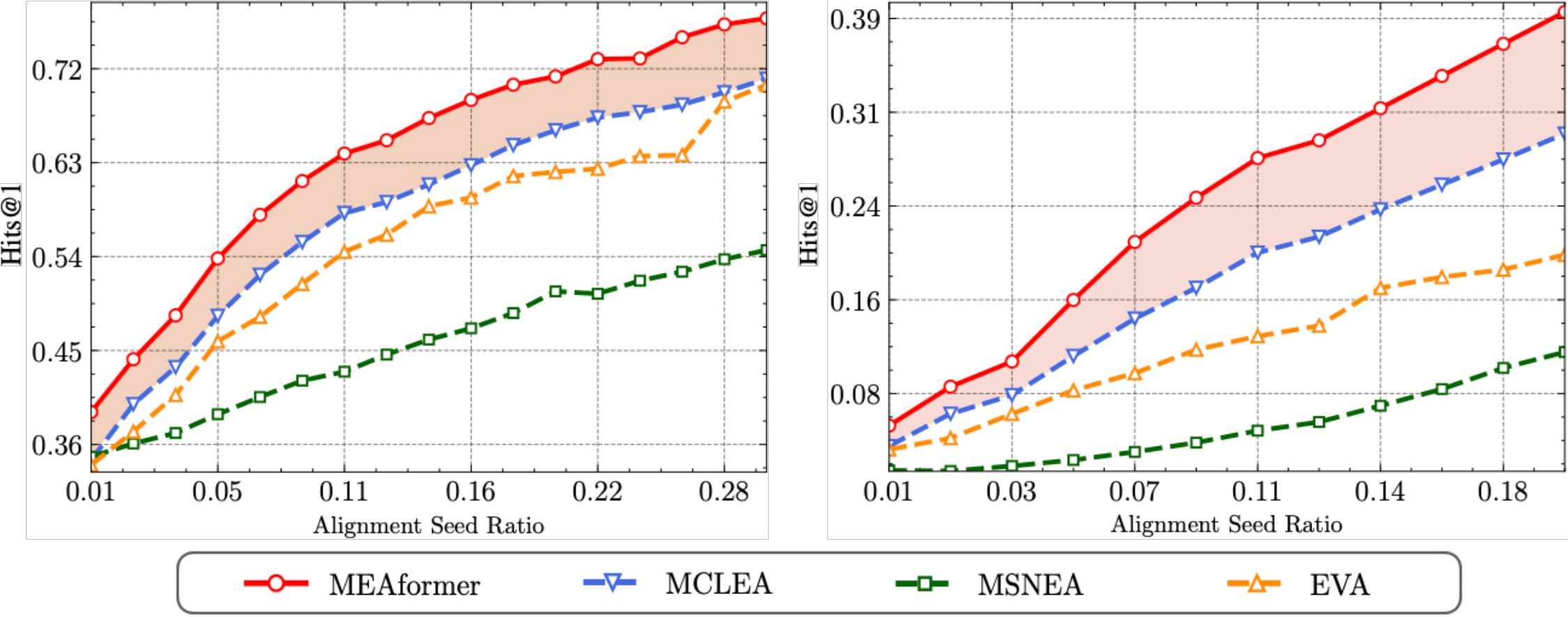}
  \vspace{-2pt}
  \caption{\textbf{Low Resource.} Models' Hits$@$1 performance with fewer seed alignments on DPB15K$_{FR\text{-}EN}$ (left) and FBDB15K (right). Please see appendix for more details.}
 \vspace{-4pt}
  \label{fig:few}
\end{figure}

To discuss the model's stability with fewer seed alignments, we evaluate MEAformer on two datasets with seed alignment ratio ($R_{sa}$) ranging from 0.01 to 0.20 in  FBDB15K and from 0.01 to 0.30 in  DBP15K$_{FR-EN}$. 
Figure \ref{fig:few} presents a clear gap that persists as the ratio increases, which becomes larger when $R_{sa}$ exceeds about 0.10.
It is worth noting that MEAformer can even achieve $.392$ on Hits$@$1 in DBP15K$_{FR\text{-}EN}$ with only $1\%$ seed alignments, compared with $.344$ for MCLEA, showing its potential for few-shot EA, which will be further explored in our future works.

\subsubsection{\textbf{Component Analysis.}}
\begin{figure}[!htbp]
\vspace{-1pt}
  \centering
  \includegraphics[trim=11 0 0 0, width=0.92\linewidth]{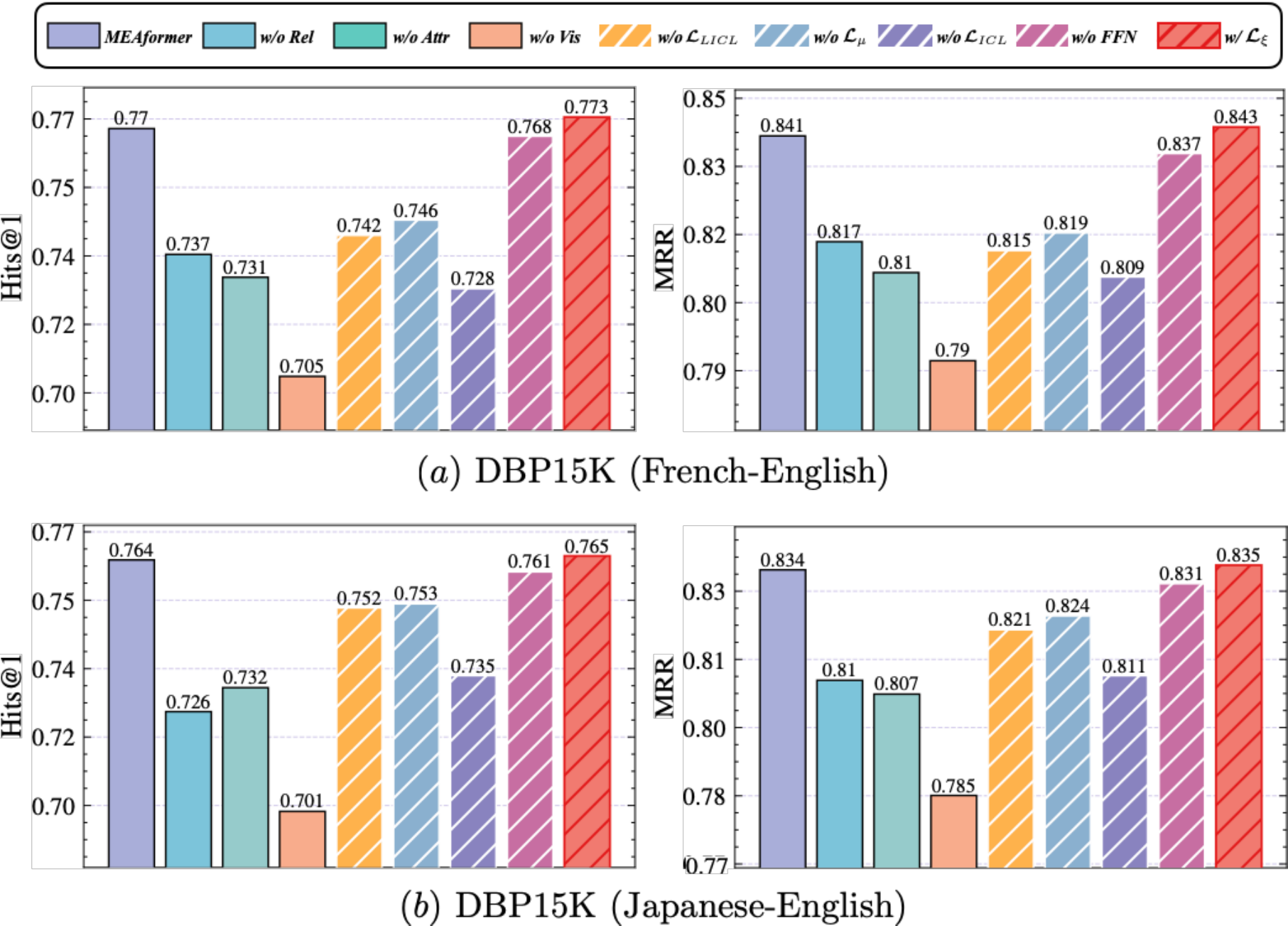}
  \vspace{-2pt}
  \caption{\textbf{Component Analysis} for MEAformer on DBP15K$_{FR\text{-}EN}$ (up) and    DBP15K$_{JA\text{-}EN}$ (down).}
 \vspace{-4pt}
  \label{fig:ablation}
\end{figure}

We evaluate various stripped-down versions of MEAformer in Figure \ref{fig:ablation} to present the Hits$@$1/MRR gains brought by different components.
Concretely, we find that removing the content from any modality always results in a noticeable performance decline,  especially the visual modality.
We postulate that the attribute and relation modalities may exhibit a degree of information complementarity, 
e.g., as demonstrated by the first case in Figure \ref{fig:Introcase}, the terms \textit{clubs} and \textit{national\_team} both appear in the $\mathcal{A}$ and $\mathcal{R}$ set of the entity \textit{Mario~Gomez}, indicating partial redundancy.
Besides, the training objectives, including $\mathcal{L}_{ICL}$, $\mathcal{L}_{LICL}$ and $\mathcal{L}_{\mu}$, all clearly boost performance, thereby validating the effectiveness of our training strategy.
Note that an additional objective $\mathcal{L}_{\xi}$, which is $\mathcal{L}_{\mu}$'s variant with late fusion embedding as the entity feature in \eqref{eq:cat}, can slightly enhance our model. 
We conjecture that the reason for $\mathcal{L}_{\xi}$'s relatively minor   enhancement comes from the mutual modality iteration in MHCA, leading to a reduced contribution from late fusion to the overall training process.
This hypothesis is supported by the more significant performance degradation observed when removing $\mathcal{L}_{ICL}$ as compared to removing $\mathcal{L}_{LICL}$ (i.e., the former has a greater impact).
Thus we simply discard $\mathcal{L}_{\xi}$ to optimize model efficiency. 
Finally, we posit that the FFN plays a crucial role in bilingual DBP15K datasets, but we do not observe the same level of performance improvement in monolingual datasets, which typically contain fewer types of relations and attributes. The Parameter Scaling Ability is dicussed in appendix.

\subsection{Detailed Analysis} \label{sec:analysis}
\subsubsection{\textbf{Error Analysis.}} 
To discuss MEAformer's robustness, we compare it against EVA by studying their overall prediction distributions across four evaluations (with varying random seeds) within the DBP15K dataset.
Concretely, an entity is defined as unanimously correct ($e_{uc}$) if it aligns successfully across all evaluations.
We observe that the Hits$@$1 scores for {$e_{uc}$} ($.762/.751/.761$) with respect to MEAformer are similar to the original results ($.771/.764/.770$) across three datasets. In contrast, for EVA, the scores plummet from ($.680/.673/.683$) to ($.490/.531/.575$), thus implying that MEAformer demonstrates a higher degree of stability than EVA.

\begin{figure}[!htbp]
\vspace{-1pt}
  \centering
  \includegraphics[trim=22 0 4 0, width=1.0\linewidth]{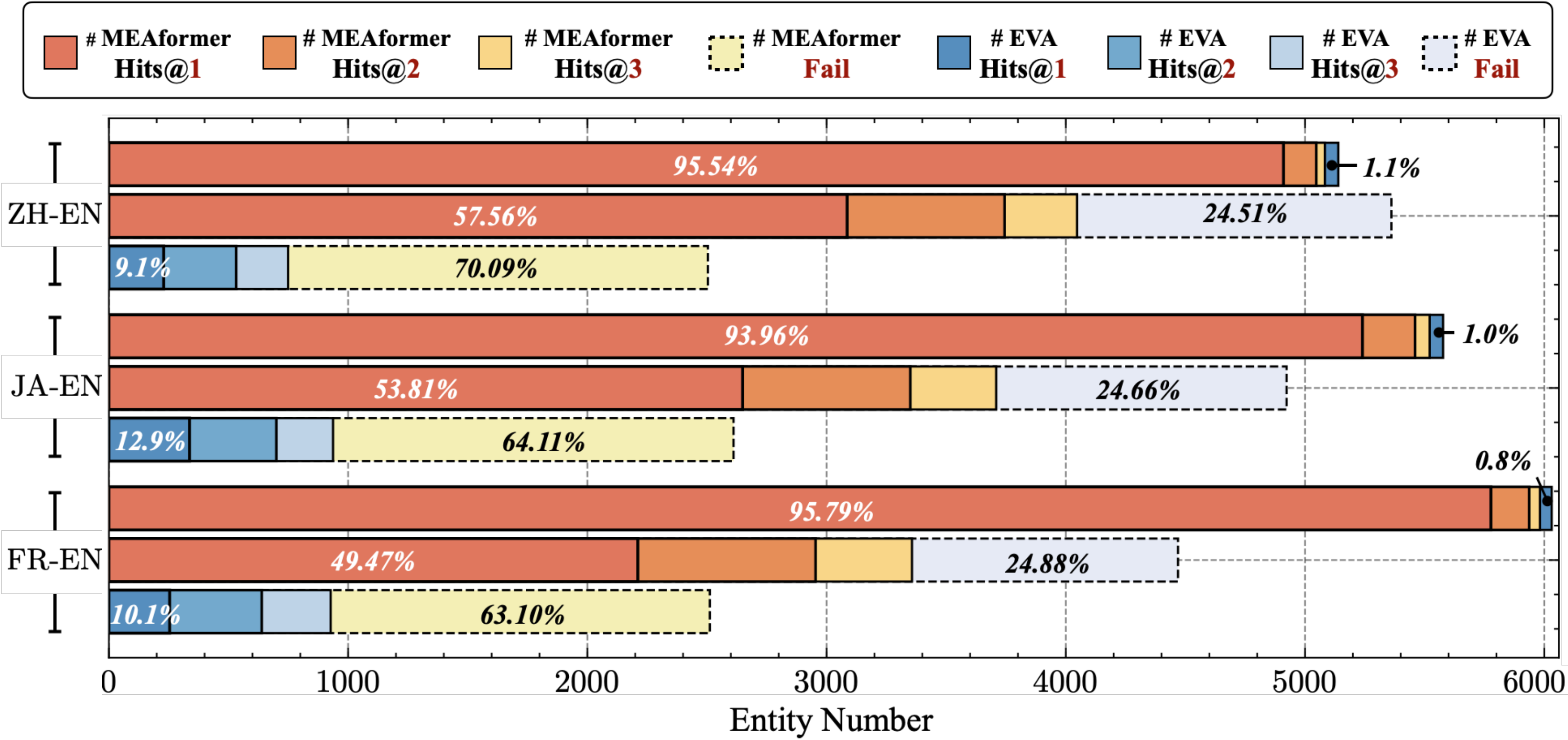}
  \vspace{-8pt}
  \caption{\textbf{Error Analysis}, where ``$\#$ Hits$@$N'' refers to the number of entities $e_{uc}$ ranked in model's top N list,  and ``$\#$ Fail'' refers to the number of  $e_{uc}$ that model failed in.}
  \vspace{-6pt}
  \label{fig:error}
\end{figure}

In Figure \ref{fig:error}, we display the intersection of model predictions to compare the capacity of each model for mutual error correction. Our findings indicate that MEAformer accurately predicts ($57.56\%/53.81\%/49.47\%$) of the entities that EVA incorrectly predicts. Conversely, EVA manages to predict just ($9.15\%$/ $12.91\%$/$10.11\%$) of the entities that MEAformer fails to align.
Also, for those entities predicted correctly by EVA, MEAformer successfully aligns a significant majority ($98.91\%/99.00\%/99.19\%$) of them at Hits$@$3. 

\subsubsection{\textbf{Modality Distribution.}} 
\begin{figure}[!htbp]
\vspace{-1pt}
  \centering
  \includegraphics[width=1.0\linewidth]{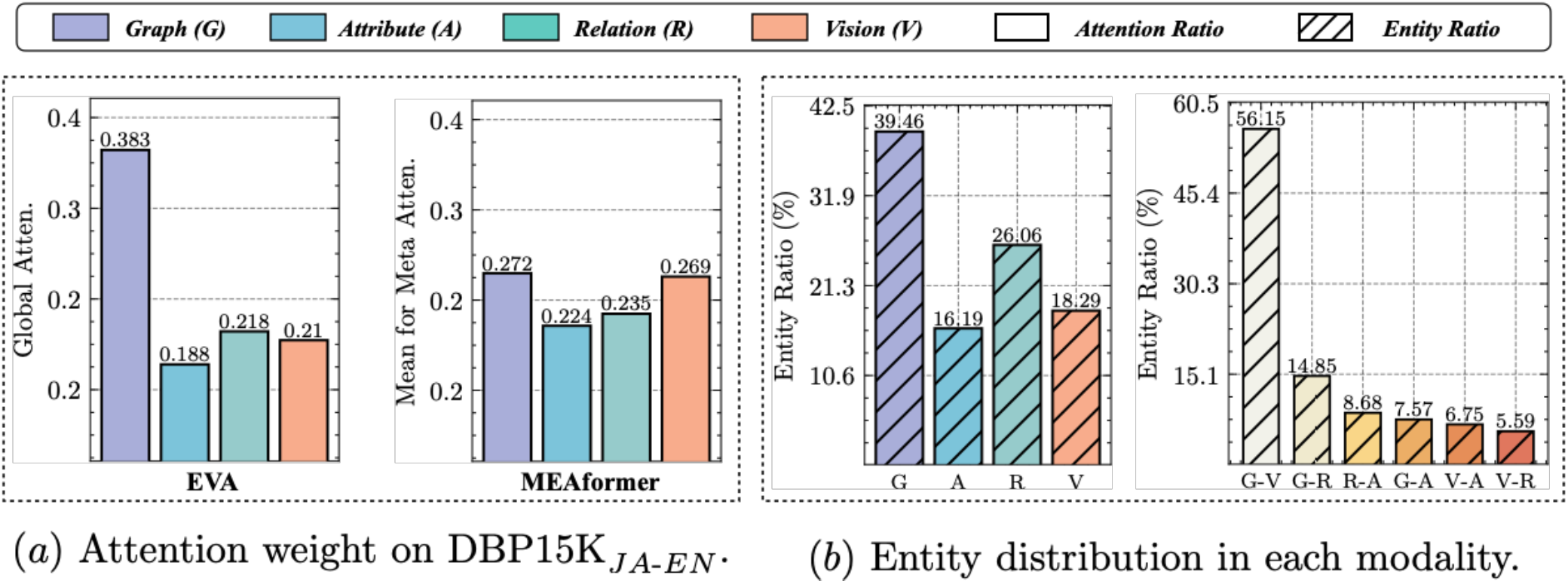}
  \vspace{-8pt}
  \caption{\textbf{Modality Distribution} for EVA and MEAformer.}
  \vspace{-6pt}
  \label{fig:modal-analysis}
\end{figure}

In Figure \ref{fig:modal-analysis}(a), we analyze the distribution for modality weights.
EVA's constant weight distribution suggests that the graph modality is extremely prominent while the weights of the other modalities are relatively average and low.
In contrast, the overall average distribution of our generated meta modality weights appears more uniform. 
Furthermore, in Figure \ref{fig:modal-analysis}(b), 
we record the entity distribution of the datasets for independent or combined modality weight preference (i.e., reach the highest proportion in each modality).
We observe that although our average weights are evenly distributed, a distinct modality preference distribution for different entities still exists, resembling EVA's modality weight. 
This partly 
supports our motivations for modality adaptation, where the global (KG-level) and local (entity-level) weight optimum are simultaneously achieved.

\subsubsection{\textbf{Case Analysis.}} 
Several representative examples are shown in Figure \ref{fig:Introcase}.
We find those {\emph{(sports) athletes}} and {\emph{(film) stars}} generally have a stronger overall reliance on the vision and graph (structure) modalities, as their appearance characteristics and visual backgrounds are typically more recognizable.
Besides, {\emph{military}} or {\emph{political}} related entities are often highly dependent on the relation modality. We owe this to the fact that the relationships among such entities are usually more complex and diverse.

Moreover, the noise involved in certain modalities may occasionally impede model judgment, as evidenced by the \textit{arena lviv} example  in the third case where the lawn visual pattern provides limited assistance, and in the second case,  where multiple dissimilar images of air force emblem refer to the same entity. These instances are commonly observed in real-world internet multi-modal data due to the abstraction of single-modal content, e.g., images of countries can be either flags or unique landmarks, and images of locations can be photos or maps.

\begin{table}[!htbp]
  \centering
\centering
{
\vspace{-5pt}
\caption{Part of the representative examples whose meta weights of corresponding modalities rank relatively high.
}
\label{tab:case}
}
\vspace{-2pt}
\resizebox{1.0\linewidth}{!}{
\begin{tabular}{c|l}
\hline
 & \\ [-2ex]
\textbf{Modality} & \makebox[8.3cm][c]{\textbf{Examples}} \\
 & \\ [-2ex]
\hline
 & \\ [-2ex]
\multirow{3}*{Vision}
&\texttt{\small Rivaldo, Reinaldo_da_Cruz_Oliveira, Gilberto_Silva, }\\
&\texttt{\small Didier_Zokora, Kate_Bush, Billie_Jean, Take_My_Breath_Away,}  \\
&\texttt{\small Léon_Blum, Celine_Dion, Ophiuchus, Lacerta, Costinha}  \\
& \\ [-2ex]
\hline
 & \\ [-2ex]
\multirow{3}*{Attribute}
&\texttt{\small Kingdom_of_England, Roman Republic, Timurid_Empire,}\\
&\texttt{\small 2014_AFC_Cup, 2015–16_Football_League_Cup, Kuwait_City,}  \\
&\texttt{\small Calabria, Hong_Kong, Amsterdam, Hamburg, Kiev}  \\
& \\ [-2ex]
\hline
 & \\ [-2ex]
\multirow{3}*{Relation}
&\texttt{\small United_States, France, Michael_Jackson, The_Beatles, }\\
&\texttt{\small Sony_Mobile, Microsoft, Apple_Inc, Queen_Victoria, Akihito,}  \\
&\texttt{\small European_People's_Party,English_language, Aramaic_alphabet}  \\
& \\ [-2ex]
\hline
 & \\ [-2ex]
\multirow{3}*{Graph}
&\texttt{\small Vágner_Love, David_Villa, Jimmy_Bullard, Torsten_Frings,}\\ 
&\texttt{\small Hydra_(constellation), Ursa_Major, Montricher, Delphinus,}  \\
&\texttt{\small Scorpius, Volans, Locarno, Crissier, Schmiedrued, Tramelan}  \\
[-2ex] \\
\hline
\end{tabular}}
\vspace{-4pt}
\end{table}
Table \ref{tab:case} records detailed entity examples from the DBP15K that have extremely unbalanced meta weights and achieve high scores in independent modalities. 
We observe that those entities with high confidence in graph  modality are KG-specific, and manually summarize several patterns and trends as follows:
\begin{itemize}
    \item Football players (e.g., \textit{David$\_$Villa} ) $-$ DBP15K$_{ZH\text{-}EN}$.
    \item Constellations (e.g., \textit{Ursa$\_$Major} ) $-$ DBP15K$_{JA\text{-}EN}$.  
    \item Switzerland's areas (e.g., \textit{Crissier} ) $-$ DBP15K$_{FR\text{-}EN}$.
\end{itemize}
Compared to other entities, we find that these examples tend to exhibit the following characteristics:
\textbf{\textit{(\rmnum{1})}} Larger node degree (i.e., the number of edges linked to the entity node). 
\textbf{\textit{(\rmnum{2})}} Limited relation types.
\textbf{\textit{(\rmnum{3})}} Associated entities predominantly belonging to the same type or the same level within the ontology \cite{DBLP:conf/www/GengC0PYYJC21} (i.e., the graph is similar to a non-hierarchical social network).

\vspace{-2pt}
\section{Conclusion and Future Directions}
In this research, we have presented a novel strategy, termed \textit{Meta Modality Hybrid}, for multi-modal entity alignment in Knowledge Graphs, facilitating the development of adaptive modality preferences. We also introduced a model-agnostic strategy, \textit{Modal-Aware Hard Entity Replay}, designed to amplify the model's sensitivity to subtle details. 
We believe that our findings and proposed method MEAformer can provide valuable insights for future research in multi-modal entity alignment.

Moreover, we noticed that the frequent absence of visual modality in Multi-Modal Entity Alignment (MMEA) is a prevalent challenge. It affects the alignment's robustness and process. In our future work, we will tackle this issue and improve our handling of missing visual modality, potentially boosting alignment models' performance and scope.

\begin{acks}
This work was supported by the National Natural Science Foundation of China (NSFCU19B2027/NSFC91846204), joint project DH-2022ZY0012 from Donghai Lab, and the EPSRC project ConCur (EP/V050869/1).
\end{acks}

\bibliographystyle{ACM-Reference-Format}
\bibliography{main}

\appendix
\clearpage

\appendix
\section{Appendix}
\begin{table*}[!htbp]
    \centering
    \caption{Statistics for datasets, where ``EA pairs'' refers to the pre-aligned entity pairs. Note that not all entities have the associated images or the equivalent counterparts in the other KG.}
    \label{tab:dataset}
    \vspace{-3pt}
    \renewcommand\arraystretch{1.0}
    \resizebox{0.73\linewidth}{!}{
    \begin{tabular}{@{}l|c|cccccccc@{}}
        \toprule
        \makebox[2.5cm][c]{Dataset} & KG & \# Ent. & \# Rel. & \# Attr. & \# Rel. Triples & \# Attr. Triples & \# Image & \# EA pairs \\
        \midrule
        \multirow{2}*{DBP15K$_{ZH-EN}$} & ZH {\footnotesize (Chinese)} & 19,388 & 1,701 & 8,111 & 70,414 & 248,035 & 15,912 & \multirow{2}*{15,000} \\
        & EN {\footnotesize (English)} & 19,572 & 1,323 & 7,173 & 95,142 & 343,218 & 14,125 \\
        \midrule
        \multirow{2}*{DBP15K$_{JA-EN}$} & JA {\footnotesize (Japanese)} & 19,814 & 1,299 & 5,882 & 77,214 & 248,991 & 12,739 & \multirow{2}*{15,000} \\
        & EN {\footnotesize (English)} & 19,780 & 1,153 & 6,066 & 93,484 & 320,616 & 13,741 \\
        \midrule
        \multirow{2}*{DBP15K$_{FR-EN}$} & FR {\footnotesize (French)} & 19,661 & 903 & 4,547 & 105,998 & 273,825 & 14,174 & \multirow{2}*{15,000} \\
        & EN {\footnotesize (English)} & 19,993 & 1,208 & 6,422 & 115,722 & 351,094 & 13,858 \\
        \midrule
        \multirow{2}*{FBDB15K} & FB15K & 14,951 & 1,345 & 116 & 592,213 & 29,395 & 13,444 & \multirow{2}*{12,846} \\
        & DB15K & 12,842 & 279 & 225 & 89,197 & 48,080 & 12,837 \\
        \midrule
        \multirow{2}*{FBYG15K} & FB15K & 14,951 & 1,345 & 116 & 592,213 & 29,395 & 13,444 & \multirow{2}*{11,199} \\
        & YAGO15K & 15,404 & 32 & 7 & 122,886 & 23,532 & 11,194 \\
        \bottomrule
    \end{tabular}
    }
    \vspace{-3pt}
    \end{table*}

\subsection{Dataset Statistics}
Our detailed dataset statistics are presented in Table \ref{tab:dataset}, which are consistent with \cite{DBLP:conf/coling/LinZWSW022}.
Note that 
a set of pre-aligned entity pairs are offered for guidance,  
which is proportionally split into a training set (seed alignments $\mathcal{S}$) and a testing set $\mathcal{S}_{te}$ based on the given seed alignment ratio ($R_{sa}$).

\subsection{Supplementary for Baselines}
\subsubsection{\textbf{Attribute Value.}}
Those attribute triples $<$\textit{entity}, \textit{attribute}, \textit{value}$>$ in KGs  have been researched in many previous EA works \cite{DBLP:conf/aaai/TrisedyaQZ19,DBLP:conf/emnlp/LiuCPLC20,DBLP:conf/ijcai/Tang0C00L20,DBLP:conf/kdd/ChenL00WYC22,DBLP:conf/icde/ZhongZFD22}.
Nevertheless, in order to focus on our key subject,
we do not utilize the contents of \textit{value} parts in this work which are mainly string formats like specific date, land area or coordinate position.
Moreover, our MEAformer still outperforms those value-aware methods when the vision modality is involved.

\subsubsection{\textbf{PLM-based Co-training.}}
As shown in Table \ref{tab:PLM-overall}, our model also outperforms those PLM-based co-training EA methods that fine-tune the PLMs (e.g., BERT \cite{DBLP:conf/naacl/DevlinCLT19}) during or before model training.
Note that their learnable parameters ($>$ 110M) are much larger than MEAformer (14M). 
Those baseline results are excerpted from \cite{DBLP:conf/icde/ZhongZFD22} where the BERT-INT \cite{DBLP:conf/ijcai/Tang0C00L20} is reproduced with entity descriptions removed. 
Note that the attribute values are involved in those two algorithms but absent in ours.

\subsubsection{\textbf{Baseline Analysis.}}
We owe the lower performance of translation based methods (e.g., MSNEA) to their reliance on semantics assumptions, which limits their ability to capture the complex structural information among entities for alignment.
Some works \cite{DBLP:conf/ijcai/WuLF0Y019,DBLP:conf/cikm/YangWZQWHH21} assert that 
the structural information plays an important role in the EA task. 
By performing graph convolution over an entity’s neighbors, GCN is able to involve more structural characteristics of knowledge graphs, while the translation assumption in translation-based models focuses more on the relationship among heads, tails and relations.

\subsubsection{\textbf{Scaling Analysis.}}
In the light-gray shaded region of Table \ref{tab:scaling}, we further discuss the impact of the Transformer architecture. 
Remarkably, even when the hidden layer dimension $d$ is reduced to 64 (parameter count reduced to 2.9M), MEAformer maintains superior performance compared to all baselines.  For comparison, we modified our MMH architecture by employing the conventional inter-modal attention mechanism for instance-level modal preference weighting, as opposed to our original MHCA module. The findings reveal that MEAformer achieves the optimal performance with the same parameter count, and our model's advantage increases as the parameter count decreases. This affirms the significance of our meta modality hybrid mechanism in MMEA and underscores the necessity of the MHCA module's design.

\begin{table*}[ht]
    \centering
	\tabcolsep=0.3cm
    \renewcommand\arraystretch{1.0}
        \caption{\textbf{Scaling Analysis:} Non-iterative results  without (w/o) surface forms (SF) on three \textbf{bilingual} datasets.
    }
    \label{tab:scaling}
    \vspace{-2pt}
    \resizebox{0.80\linewidth}{!}{
    \begin{tabular}{@{}l|l|c|ccc|ccc|ccc}
        \toprule
        & \multirow{2}*{\makebox[3.4cm][c]{Models}} & \multirow{2}*{Para.} & \multicolumn{3}{c|}{DBP15K$_{ZH-EN}$} & \multicolumn{3}{c|}{DBP15K$_{JA-EN}$} & \multicolumn{3}{c}{DBP15K$_{FR-EN}$} \\
        & & & {\scriptsize H@1} & {\scriptsize H@10} & {\scriptsize MRR} & {\scriptsize H@1} & {\scriptsize H@10} & {\scriptsize MRR} & {\scriptsize H@1} & {\scriptsize H@10} & {\scriptsize MRR} \\
        \midrule
        \parbox[t]{2mm}{\multirow{8}{*}{\rotatebox[origin=c]{90}{w/o SF}}} 
        & EVA* {\footnotesize \cite{DBLP:conf/aaai/0001CRC21}} & 13.3M &
        {.680} & {.910} & {.762} & {.673} & {.908} & {.757} & {.683} & \underline{.923} & {.767} \\
        & MSNEA* {\footnotesize {\cite{DBLP:conf/kdd/ChenL00WYC22}}} & 14.1M & .601 & .830 & .684 & .535 & .775 & .617 & .543 & .801 & .630 \\
        & MCLEA* {\footnotesize {\cite{DBLP:conf/coling/LinZWSW022}}} & 13.2M &
        \underline{.715} & \underline{.923} & \underline{.788} & \underline{.715} & \underline{.909} & \underline{.785} & \underline{.711} & {.909} & \underline{.782} \\
        & \CC\textbf{MEAformer} {\footnotesize (Ours)} & \CC13.7M &
        \CC\textbf{.771} & \CC\textbf{.951} & \CC\textbf{.835} & \CC\textbf{.764} & \CC\textbf{.959} & \CC\textbf{.834} & \CC\textbf{.770} & \CC\textbf{.961} & \CC\textbf{.841} \\
        & \CC{- MEAformer} {\footnotesize ($d=128$)} & \CC5.8M &
        \CC{.757} & \CC{.939} & \CC{.822} & \CC{.755} & \CC{.951} & \CC{.826} & \CC{.758} & \CC{.957} & \CC{.830} \\
        & \CC{- MEAformer} {\footnotesize ($d=64$)} & \CC\textbf{2.9M} &
        \CC{.731} & \CC{.929} & \CC{.802} & \CC{.726} & \CC{.936} & \CC{.800} & \CC{.727} & \CC{.944} & \CC{.803} \\
        & \CC{- Atten-based MMH} {\footnotesize ($d=300$)} & \CC13.3M &
        \CC{.762} & \CC{.945} & \CC{.827} & \CC{.756} & \CC{.941} & \CC{.823} & \CC{.751} & \CC{.943} & \CC{.822} \\
        & \CC{- Atten-based MMH} {\footnotesize ($d=128$)} & \CC5.6M &
        \CC{.741} & \CC{.931} & \CC{.809} & \CC{.732} & \CC{.928} & \CC{.803} & \CC{.726} & \CC{.921} & \CC{.799} \\
        \bottomrule
    \end{tabular}
    \vspace{-4pt}
    }
\end{table*}

\begin{table*}[ht]
    \centering
	\tabcolsep=0.3cm
    \renewcommand\arraystretch{1.0}
    \caption{Results of our \textbf{non-iterative} models on three \textbf{bilingual} datasets compared with those \textbf{PLM-based} co-training or fine-tuning methods. Note that the entity descriptions are not available in all datasets for fairness.  
    }
    \label{tab:PLM-overall}
    \vspace{-2pt}
    \resizebox{0.81\linewidth}{!}{
    \begin{tabular}{@{}l|l|c|ccc|ccc|ccc}
        \toprule
        & \multirow{2}*{\makebox[3.85cm][c]{Models}} & \multirow{2}*{Para.} & \multicolumn{3}{c|}{DBP15K$_{ZH-EN}$} & \multicolumn{3}{c|}{DBP15K$_{JA-EN}$} & \multicolumn{3}{c}{DBP15K$_{FR-EN}$} \\
        & & & {\scriptsize H@1} & {\scriptsize H@10} & {\scriptsize MRR} & {\scriptsize H@1} & {\scriptsize H@10} & {\scriptsize MRR} & {\scriptsize H@1} & {\scriptsize H@10} & {\scriptsize MRR} \\
        \midrule
        \parbox[t]{2mm}{\multirow{3}{*}{\rotatebox[origin=c]{90}{w/ SF}}} 

        & BERT-INT {\footnotesize {\cite{DBLP:conf/ijcai/Tang0C00L20}}} & $>$110M &
        {.814} & {.837} & {.820} & {.806} & {.835} & {.820} & \underline{.987} & {.992} & \underline{.990} \\
        & SDEA {\footnotesize {\cite{DBLP:conf/icde/ZhongZFD22}}} & $>$110M & \underline{.870} & \underline{.966} & \underline{.910} & \underline{.848} & \underline{.952} & \underline{.890} & .969 & \underline{.995} & .980 \\
        & \CC\textbf{MEAformer} {\footnotesize (Ours)} & \CC\textbf{14.2M} &
        \CC\textbf{.948} & \CC\textbf{.993} & \CC\textbf{.965} & \CC\textbf{.977} & \CC\textbf{.999} & \CC\textbf{.986} & \CC\textbf{.991} & \CC\textbf{1.00} & \CC\textbf{.995} \\
        \bottomrule
    \end{tabular}
    }
    \vspace{-3pt}
\end{table*}

\subsection{Efficiency Analysis}
Specifically, we set the total epochs for each model into $3000$ with early stopping rule adopted to get the {\emph{best achievable}} performances in Figure \ref{fig:effic} (right), because a larger number of epochs will make the learning rate change smoother with the warm-up schedule equipped.
We observe that this setting help further increase our performances on FBDB15K ($R_{sa}$ $=$ $0.2$) from $.417/.518$ to $.431/.530$ (Hits$@$1 / MRR), and increase our performances on DBP15K$_{ZH\text{-}EN}$  from $.771/.835$ to $.773/.837$. Note that we only record the convergence curve before overfitting in Figure \ref{fig:effic} (left).  

\subsection{Low Resource Supplementary}
\begin{figure}[!htbp]
  \centering
  \includegraphics[trim=18 0 0 0, width=0.95\linewidth]{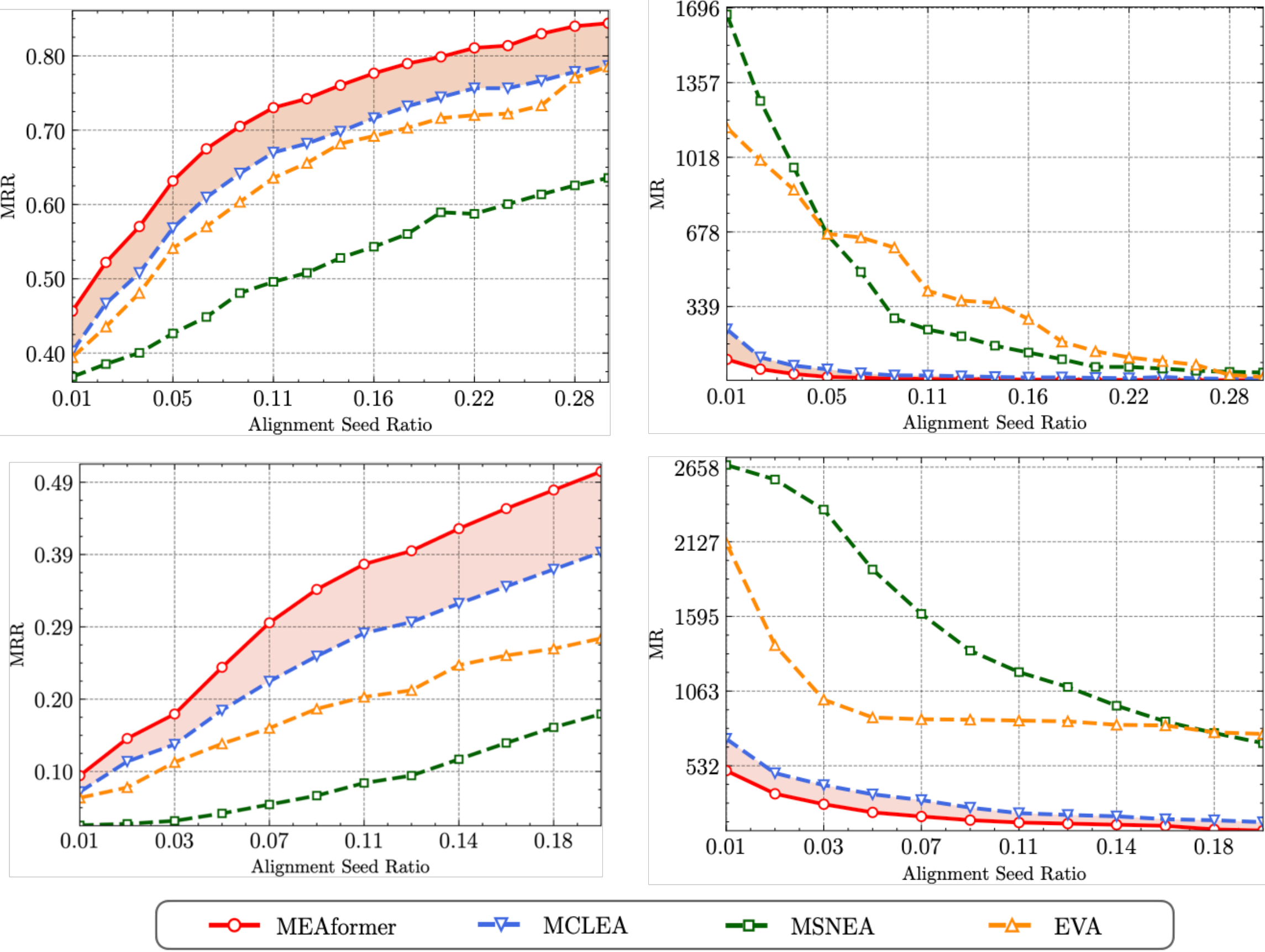}
  \vspace{-4pt}
  \caption{ Models' MRR$\uparrow$ / MR$\downarrow$ performance with fewer seed alignments on DPB15K$_{FR\text{-}EN}$ (up) and FBDB15K (down).}
  \label{fig:fewsup}
  \vspace{-7pt}
\end{figure}

We evaluate MEAformer with seed alignment ratio ($R_{sa}$) \emph{\{0.01, 0.02, 0.03, 0.05, 0.07, 0.09, 0.11, 0.12, 0.14, 0.16, 0.18, 0.20\}} in  FBDB15K and  \emph{\{0.01, 0.02, 0.03, 0.05, 0.07, 0.09, 0.11, 0.12, 0.14, 0.16, 0.18, 0.20, 0.22, 0.24, 0.26, 0.28, 0.30\}} in  DBP15K$_{FR-EN}$.
The Hits$@$1 results are shown in Figure \ref{fig:few}, and we present the MRR / MR results in Figure \ref{fig:fewsup}.
These trends all demonstrate the superiority of our model.

\subsection{Discussions}

{\color{red} {Q1}}: 
Why not apply the vision-language pre-training (VLP) models or combine the vision transformers (ViTs) with pure PLMs for multi-modal entity encoding and modality hybrid?

\vspace{1pt}
{\color{blue} {A1}}:
We found that using ViTs, PLMs, and VLP \cite{DBLP:journals/ftcgv/GanLLWLG22} models alone cannot effectively encode graph structures, relations, attributes, and visual images to obtain their corresponding modality features. More importantly, these models often have a large number of parameters, which will decrease the training efficiency and make it difficult to fit the {\emph{limited entity data (at the thousand level)}} from multiple modalities.

\vspace{3pt}
{\color{red} {Q2}}: 
Why does the data from different modalities and the entities in the source and target KGs share the same FFN and query / key / value matrix parameters?

\vspace{1pt}
{\color{blue} {A2}}:
To simplify the network and prevent overfitting, we refer to previous works \cite{DBLP:conf/coling/LinZWSW022,DBLP:conf/kdd/ChenL00WYC22,DBLP:conf/aaai/0001CRC21} and chose to share all model parameters of the two KGs.
Additionally, in the MMH module, the data from each modality is encoded by its own independent modality encoder before entering the DCMW block. Those encoders here play the roles of mapping multi-source data into the same embedding space, which eliminates the need to explicitly defining the modal-aware attention matrix and simplifies the training process.

%

\subsection{Model Details}
We reproduce EVA \cite{DBLP:conf/aaai/0001CRC21}, MSNEA \cite{DBLP:conf/kdd/ChenL00WYC22}, and MCLEA \cite{DBLP:conf/coling/LinZWSW022} based on their source code \footnote{\color{NavyBlue} \url{https://github.com/cambridgeltl/eva}}\footnote{{\color{NavyBlue} \url{https://github.com/lzxlin/MCLEA}}}\footnote{\color{NavyBlue} \url{https://github.com/liyichen-cly/MSNEA}} with their
original model pipelines unchanged.
We note that in the datasets FBDB15K / FBYG15K, the FFN layer of MEAformer is removed from DCMW as elaborated in Section \ref{sec:ablation}, which simplifies the model structure and slightly increase the MRR by $0.001 \sim 0.003$. 

\subsection{Metric Details}
\subsubsection{\textbf{Hits$@$N}} describes the fraction of true aligned 
 target entities that appear in the first N entities of the sorted rank list:
\begin{equation}
  \vspace{-2pt}
    \operatorname{Hits} @ \text{N}=\frac{1}{|\mathcal{S}_{te}|} \sum_{i=1}^{|\mathcal{S}_{te}|} \mathbb{I}[{\text {rank}_i} \leqslant \text{N}]\, ,
\end{equation}
where  ${{\text{rank}}_{i}}$ refers to the rank position of the first correct mapping for the i-th query entities and $\mathbb{I}=1$ if ${\text {rank}_i} \leqslant N$ and 0 otherwise.
$\mathcal{S}_{te}$ refers to the testing alignment set.
\subsubsection{\textbf{MRR}} (Mean Reciprocal Ranking $\uparrow$) is a statistic measure for evaluating many algorithms that produce a list of possible responses to a sample of queries, ordered by probability of correctness. 
In the field of EA, the reciprocal rank of a query entity (i.e., an entity from the source KG) response is the multiplicative inverse of the rank of the first correct alignment entity in the target KG.
MRR is the average of the reciprocal ranks of results for a sample of candidate alignment entities:
\begin{equation}
    \vspace{-2pt}
    \mathbf{MRR}=\frac{1}{|\mathcal{S}_{te}|} \sum_{i=1}^{|\mathcal{S}_{te}|} \frac{1}{\text {rank}_i} \,.
\end{equation}
\subsubsection{\textbf{MR}} (Mean Rank $\downarrow$) computes the arithmetic mean over all individual ranks which is similar to MRR:
\begin{equation}
    \mathbf{MR}=\frac{1}{|\mathcal{S}_{te}|} \sum_{i=1}^{|\mathcal{S}_{te}|} {\text {rank}_i} \,. 
\end{equation}
Note that MR is sensitive to any model performance changes, not just those occurring under a certain cutoff, and thus reflects the average performance.

\end{document}